\documentclass[letterpaper, 10pt, journal, twoside]{IEEEtran}

\usepackage{amssymb}
\usepackage[numbers, sort&compress]{natbib}
\usepackage{multicol}
\usepackage[usenames,dvipsnames,table]{xcolor}
\usepackage[bookmarks=true]{hyperref}
\usepackage{pdfx}
\usepackage{wrapfig}
\usepackage{booktabs}
\usepackage{xspace}
\usepackage{amsmath}
\usepackage{mathabx}
\usepackage[capitalize]{cleveref}
\usepackage{graphicx}
\usepackage{adjustbox}
\usepackage{multirow}
\usepackage{makecell}
\usepackage{caption}
\usepackage{tikz}
\usepackage{array}
\usepackage{subcaption}
\usepackage{tcolorbox}
\usepackage{stfloats}
\usepackage{enumitem}

\hypersetup{
    colorlinks=true,
    linkcolor=orange,
    filecolor=magenta,      
    urlcolor=orange,
    citecolor=orange,
    pageanchor=false,
}

\newcommand{\textgbf}[1]{\textcolor{darkgray}{\textbf{#1}}}

\newcommand{\TaxonomyName}{$\bigstar$-Gen\xspace}
\newcommand{\BenchName}{BridgeV2-$\bigstar$\xspace}

\newcommand{\graymidrule}{\arrayrulecolor{black!30}\midrule\arrayrulecolor{black}}

\definecolor{pcheck}{rgb}{0.651, 0.380, 0.102} %
\definecolor{fcheck}{rgb}{0.004, 0.522, 0.443} %

\newcommand{\fcheck}{\textcolor{fcheck}{\checkmark}}

\newcommand{\Aug}{Image Augmentations\xspace}
\newcommand{\AugShort}{V-AUG\xspace}
\newcommand{\AugShorter}{AUG\xspace}

\newcommand{\VScene}{Visual Scene\xspace}
\newcommand{\VSceneShorter}{SC\xspace}
\newcommand{\VSceneShort}{V-SC\xspace}

\newcommand{\VObject}{Visual Task Object\xspace}
\newcommand{\VObjectShort}{V-OBJ\xspace}
\newcommand{\VObjectShorter}{OBJ\xspace}

\newcommand{\View}{Viewpoint\xspace}
\newcommand{\ViewShort}{V-VIEW\xspace}
\newcommand{\ViewShorter}{VIEW\xspace}

\newcommand{\SMulti}{Multi-Object Referencing\xspace}
\newcommand{\SMultiShort}{S-MO\xspace}
\newcommand{\SMultiShorter}{MO\xspace}

\newcommand{\SProp}{Object Properties\xspace}
\newcommand{\SPropShort}{S-PROP\xspace}
\newcommand{\SPropShorter}{PROP\xspace}

\newcommand{\SInternet}{Internet Knowledge\xspace}
\newcommand{\SInternetShort}{S-INT\xspace}
\newcommand{\SInternetShorter}{INT\xspace}

\newcommand{\SRephrase}{Language Rephrase\xspace}
\newcommand{\SRephraseShort}{S-LANG\xspace}
\newcommand{\SRephraseShorter}{LANG\xspace}

\newcommand{\SAff}{Human Affordances\xspace}
\newcommand{\SAffShort}{S-AFF\xspace}
\newcommand{\SAffShorter}{AFF\xspace}

\newcommand{\BUnObj}{Hidden Object\xspace}
\newcommand{\BUnObjShort}{B-HOBJ\xspace}
\newcommand{\BUnObjShorter}{HOBJ\xspace}

\newcommand{\BUnScene}{Hidden Scene\xspace}
\newcommand{\BUnSceneShort}{B-HSC\xspace}

\newcommand{\VBPose}{Object Poses\xspace}
\newcommand{\VBPoseShort}{VB-POSE\xspace}
\newcommand{\VBPoseShorter}{POSE\xspace}

\newcommand{\VBObject}{Morphed Objects\xspace}
\newcommand{\VBObjectShort}{VB-MOBJ\xspace}
\newcommand{\VBObjectShorter}{MOBJ\xspace}

\newcommand{\VBScene}{Interacting Scene\xspace}
\newcommand{\VBSceneShort}{VB-ISC\xspace}
\newcommand{\VBSceneShorter}{ISC\xspace}

\newcommand{\VBRobot}{Robot Embodiment\xspace}
\newcommand{\VBRobotShort}{VB-ROB\xspace}
\newcommand{\VBRobotShorter}{ROB\xspace}

\newcommand{\VBSym}{Symmetry\xspace}
\newcommand{\VBSymShort}{VB-SYM\xspace}

\newcommand{\SBAdv}{Motion Adverbs\xspace}
\newcommand{\SBAdvShort}{SB-ADV\xspace}

\newcommand{\SBClause}{Spatial Multi-Object\xspace}
\newcommand{\SBClauseShort}{SB-SMO\xspace}
\newcommand{\SBClauseShorter}{SMO\xspace}

\newcommand{\SBGrounding}{Noun Grounding\xspace}
\newcommand{\SBGroundingShort}{SB-NOUN\xspace}
\newcommand{\SBGroundingShorter}{NOUN\xspace}

\newcommand{\SBAction}{Action Verbs\xspace}
\newcommand{\SBActionShort}{SB-VRB\xspace}

\newcommand{\VSObj}{New Object Property\xspace}
\newcommand{\VSObjShort}{VS-PROP\xspace}

\newcommand{\VSBObj}{New Object\xspace}
\newcommand{\VSBObjShort}{VSB-NOBJ\xspace}
\newcommand{\VSBObjShorter}{NOBJ\xspace}

\begin{document}

\title{A Taxonomy for Evaluating Generalist Robot Manipulation Policies
}
\author{Jensen Gao$^{*1}$,
Suneel Belkhale$^{*1}$,
Sudeep Dasari$^2$, 
Ashwin Balakrishna$^2$,
Dhruv Shah$^{2,3}$,
Dorsa Sadigh$^1$
\vspace{-20pt}
}%

\renewcommand*{\bibfont}{\footnotesize}

\maketitle
\markboth{IEEE ROBOTICS AND AUTOMATION LETTERS. PREPRINT VERSION. ACCEPTED JANUARY, 2026}
{Gao \MakeLowercase{\textit{et al.}}: A Taxonomy for Evaluating Generalist Robot Manipulation Policies} 

\begin{abstract}
    Machine learning for robot manipulation promises to unlock generalization to novel tasks and environments. But how should we measure the progress of these policies towards generalization? Evaluating and quantifying generalization is the Wild West of modern robotics, with each work proposing and measuring different types of generalization in their own, often difficult to reproduce settings. In this work, our goal is (1) to outline the forms of generalization we believe are important for robot manipulation in a comprehensive and fine-grained manner, and (2) to provide reproducible guidelines for measuring these notions of generalization. We first propose \TaxonomyName, a taxonomy of generalization for robot manipulation structured around visual, semantic, and behavioral generalization. Next, we instantiate \TaxonomyName with two case studies on real-world benchmarking: one based on open-source models and the Bridge V2 dataset, and another based on the bimanual ALOHA 2 platform that covers more dexterous and longer horizon tasks. Our case studies reveal many interesting insights: for example, we observe that open-source vision-language-action models often struggle with semantic generalization, despite pre-training on internet-scale language datasets. %
    We provide videos and other supplementary material at our website \url{stargen-taxonomy.github.io}.

\end{abstract}


\renewcommand{\thefootnote}{}
\footnotetext{
Manuscript received: August 21, 2025; Revised: November 28, 2025; Accepted: January 7, 2026.

This paper was recommended for publication by Editor Markus Vincze upon evaluation of the Associate Editor and Reviewers’ comments.

This work was supported by ONR N00014-22-1-2293, DARPA W911NF2210214, and NSF 1941722.

$^*$Equal contribution. Jensen Gao, Suneel Belkhale, and Dorsa Sadigh are with $^1$Stanford University {\tt\footnotesize jenseng@stanford.edu, belkhale@stanford.edu, dorsa@cs.stanford.edu}

Sudeep Dasari and Ashwin Balakrishna are with $^2$Google DeepMind.

Dhruv Shah is with $^2$Google DeepMind and $^3$Princeton University {\tt\footnotesize shahd@princeton.edu}

Digital Object Identifier (DOI): see top of this page.
}
\renewcommand{\thefootnote}{\arabic{footnote}}

\section{Introduction}
\label{sec:introduction}
\IEEEPARstart{L}{earning}-based robotics comes with the promise of broad generalization. As an example, an ambitious goal is to train a laundry-folding robot on diverse household data that can fold laundry in new homes. If trained effectively, the robot should be able to fold unseen clothing items in new settings using its extensive prior experience. However, we have yet to reach a point in robot manipulation where policies can reliably generalize in this manner. In pursuit of this vision, recent work has focused on scaling up data collection~\cite{mandlekar2018roboturk, dasari2019robonet, ebert2021bridge, jang2022bc, brohan2022rt, brohan2023rt, shafiullah2023bringing, bharadhwaj2023roboagent, walke2023bridgedata, fang2023rh20t, khazatsky2024droid, mirchandani2024robocrowd} and developing more expressive models~\cite{brohan2023rt, kim24openvla, wang2024scaling, liu2025rdtb, black2024pi_0}, following the successes of other machine learning domains.

While these advances have led to more capable policies, it is often unclear how generalist these policies truly are. Although prior work has shown various forms of generalization, such as visual robustness to distractors, or understanding novel language instructions, there is often a lack of consistency across different evaluations. Each work proposes their own forms of generalization and evaluation conditions, usually with little transparency into how they were decided upon. As a result, it has become difficult to measure progress toward real-world deployability of these policies, which has remained largely elusive, despite promising results in the literature.

To work towards more comprehensive and systematic evaluations, %
we propose \TaxonomyName (STAR-Gen) -- a \textbf{S}ystematic \textbf{T}axonomy of the \textbf{A}xes of \textbf{R}obot \textbf{Gen}eralization. We observe that policies require generalization when there are \emph{perturbations} to the policy's \emph{inputs} or required \emph{outputs}. Therefore, to ground our taxonomy, we structure \TaxonomyName based on the input and output modalities of visuo-lingual control policies: vision, language, and actions. We categorize perturbations as \textgbf{visual}, \textgbf{semantic}, and/or \textgbf{behavioral} based on how they affect these modalities. For each combination of these labels (e.g., \textgbf{visual} only, or \textgbf{visual + behavioral}), we define more granular generalization \emph{axes}. For instance, we include \emph{Object Properties} as a \textgbf{semantic} axis, which involves generalizing from ``put carrot on plate" to ``put the orange object on the plate".

\begin{figure*}[t]
    \centering
    \vspace{2pt}
    \includegraphics[width=0.93\linewidth]{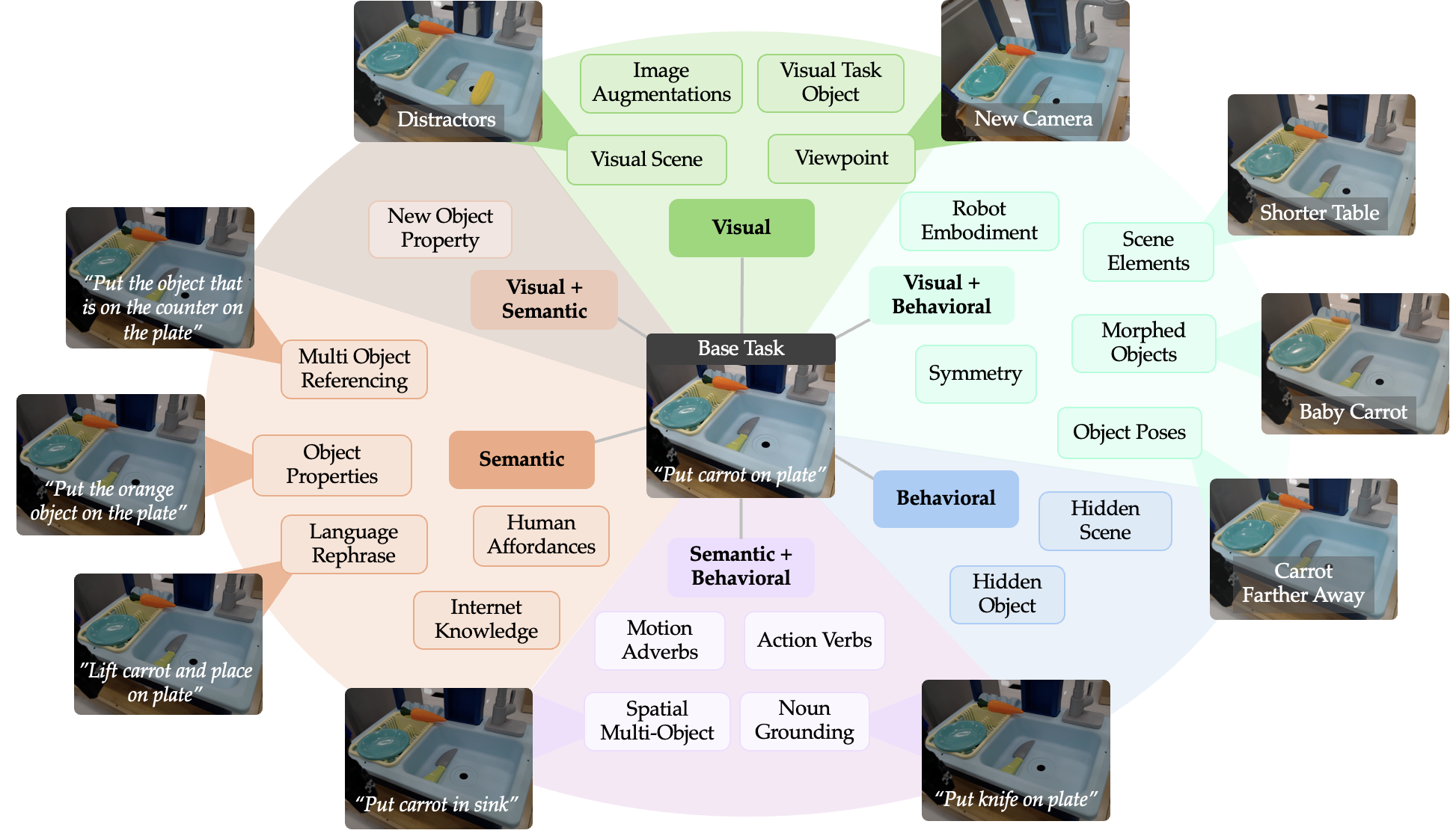}
    \vspace{6pt}
     \caption{\small Visualization of \TaxonomyName for the example base task ``put carrot on plate". \TaxonomyName is structured around perturbations to the modalities of visuo-lingual policies (\textgbf{visual}, \textgbf{semantic}, \textgbf{behavioral}), with consideration for each of their combinations, which we refer to as \emph{categories}. For each category (colored sectors), we further group perturbations into \emph{axes} (light colored boxes). We provide some example perturbations.} %
    \label{fig:axes}
    \vspace{-8pt}
\end{figure*}

To demonstrate the practical utility of \TaxonomyName, we present two real-world case studies on benchmarking generalization. First, we develop \BenchName, a benchmark based on the Bridge V2 dataset~\cite{walke2023bridgedata}, that is intended to provide a blueprint for designing generalization benchmarks using a reproducible and open-source platform. We outline the design choices of \BenchName based on our taxonomy, and use it to evaluate state-of-the-art open-source generalist manipulation policies.

Next, we use \TaxonomyName to develop an additional case study based on the bimanual ALOHA 2 platform~\cite{aldaco2024aloha} that considers more dexterous, varied, and longer-horizon tasks, supported by a large-scale real-world dataset with thousands of hours of demonstrations. This further demonstrates the broad applicability of our taxonomy to a wide range of settings.

In summary, we present \TaxonomyName, a taxonomy of generalization structured around three modalities -- vision, language, and actions -- that span the space of generalization for visuo-lingual manipulation policies. We instantiate \TaxonomyName as two real-world case studies that consist of 1600+ robot trials across 14 axes of generalization, which generate more detailed findings on state-of-the-art generalist policies and model design decisions. We hope that by guiding policy training and evaluation efforts, \TaxonomyName can help advance progress in robot manipulation.

\section{Related Work}
\label{sec:related_work}
\vspace{3pt}
To achieve broad generalization in robotics, much prior work has focused on scaling up real-world data collection. These efforts typically aim to capture diversity in both environmental conditions and task behavior~\cite{mandlekar2018roboturk, dasari2019robonet, ebert2021bridge, walke2023bridgedata, fang2023rh20t,  o2023open, khazatsky2024droid}. While these datasets are usually collected with diversity in mind, it is often unclear what forms of diversity matter, or how this diversity should be achieved. Recent works have investigated best practices for generating diverse robot data~\cite{@gao2024, lin2024data, saxena2025what}. Although these works consider various notions of diversity during data collection and corresponding axes of generalization during evaluation, these axes are neither exhaustive nor standardized.

Nevertheless, recent works have attempted to leverage these datasets for learning generalist robot policies. These works involve training large-scale, visuo-lingual policies on this data, with the goal of generalizing to a wide variety of scenarios~\cite{brohan2022rt, brohan2023rt, bharadhwaj2023roboagent, o2023open, kim24openvla, wang2024scaling, liu2025rdtb, black2024pi_0}. However, each work designs their own evaluations that often focus on a relatively narrow selection of generalization, making it challenging to assess how models make progress towards different forms of generalization.

A compelling alternative is to benchmark generalization in simulation. There has been extensive work in simulated robot manipulation platforms that support task and scene diversity~\cite{james2020rlbench, mu2021maniskill, szot2021habitat, ehsani2021manipulathor, li2023behavior, nasiriany2024robocasa}. However, these largely do not come with benchmarks that measure precise notions of generalization. While some works have studied specific distribution shifts in simulation~\cite{xing2021kitchenshift, mees2022calvin, xie2024decomposing, pumacay2024colosseum, li2024evaluating}, the generalization axes considered are usually inconsistent across works, similar to real-world efforts.

To help unify and provide structure to notions of generalization in robot manipulation, we propose \TaxonomyName, which considers a superset of generalization axes from prior works. We hope this taxonomy can be useful for developing better datasets and models that make progress towards generalization, and developing better benchmarks to capture this progress.

\section{What is Generalization?}
\label{sec:preliminaries}
In this section, we outline our preliminaries, provide a formal characterization of generalization for robot policies, and list some additional assumptions, which we will use later in \cref{sec:axes:tax} to design our taxonomy \TaxonomyName.

\subsection{Preliminaries}
\smallskip \noindent \textbf{Environment.} We define an environment as the tuple $E = (\mathcal{S}, \mathcal{O}, \mathcal{A}, \mathcal{L}, f_o, f_t)$, where $\mathcal{S}$ is the state space,  $\mathcal{O}$ is the observation space derived from $\mathcal{S}$ through observation function $f_o: \mathcal{S} \to \mathcal{O}$, $\mathcal{A}$ is the action space, and $f_t: \mathcal{S} \times \mathcal{A} \to \mathcal{S}$ is the transition function. We assume $\mathcal{O}$ consists of third-person images of a scene, and $\mathcal{A}$ consists of robot actions.

\smallskip \noindent \textbf{Task.} For an environment $E$, we define a task space $T$. A task $\tau \in T$ is defined as $\tau = (p_\tau(s_0), l_\tau, R_\tau)$, where $p_\tau(s_0)$ is an initial state distribution for $E$, $l_\tau \in \mathcal{L}$ is a language instruction, and $R_\tau : (\mathcal{S} \times \mathcal{A})^* \to \{0, 1\}$ is a success function that maps a state/action sequence to a success indicator. $p_\tau(s_0)$ defines an initial observation distribution $p_\tau(o_0)$ induced by $f_o$.

\smallskip \noindent \textbf{Policy.} A policy $\pi(a \mid o^n, l)$ takes in $n \geq 1$ observations, a language instruction, and outputs an action distribution. We define an \textbf{expert policy} $\pi_E(a \mid o^n, l)$ that produces successful episodes (where success is defined by $R_\tau$) for a given task $\tau$.

\subsection{Defining Generalization for Visuo-Lingual Policies}

\smallskip 
Generalization in robotics is often considered as the performance of a policy $\pi$ on a task $\tau'$ outside its training distribution. %
To deploy policies in diverse settings, there is a vast space of potential tasks $\tau'$ to consider. Furthermore, it can be challenging to characterize how tasks represent generalization from large datasets. %
To address these challenges and provide a theoretically grounded framework, we propose structuring our taxonomy of generalization around \emph{perturbations} of a given base task, and how they affect the core input and output modalities of a robot policy, which we formalize as follows:

\smallskip \noindent \textbf{Base Task.} A base task $\tau_B$ is a task where an end application desires a policy to perform the task (e.g., chopping a specific onion) and perturbations of it (e.g., chopping other onions).

\renewcommand{\arraystretch}{0.9}
\setlength{\tabcolsep}{2.8pt}
\begin{table*}[!ht]
\vspace{3pt}
\centering
\fontsize{6.5}{7}\selectfont
\caption{\TaxonomyName: Axes of Generalization}
\label{tab:axes}
\setlength{\extrarowheight}{2pt}
\begin{tabular}{p{0.16\linewidth}p{0.08\linewidth}p{0.43\linewidth}p{0.27\linewidth}}
\toprule
\textbf{Axis} & \textbf{Name} & \textbf{Description} & \textbf{Example Factors} \\

\rowcolor{OliveGreen!20} \specialrule{1pt}{2pt}{2pt}
\multicolumn{4}{c}{\small Visual} \\[1pt]

\textbf{\Aug} & \AugShort & Realistic generic augmentations in image space. & lighting, image blur, image contrast \\
\graymidrule
\textbf{\VScene} & \VSceneShort & Visual changes to scene elements that do not affect behavior. & surface color/texture, distractor object appearance/placement \\
\graymidrule
\textbf{\VObject} & \VObjectShort & Visual changes to task-relevant objects that do not affect behavior. & manipulated object color, other task-relevant object color (e.g., container an object is placed in) \\
\graymidrule
\textbf{\View} & \ViewShort & Changes to camera viewpoints. & camera pose, partial occlusion \\

\rowcolor{Apricot!40} \specialrule{1pt}{2pt}{2pt}
\multicolumn{4}{c}{\small Semantic} \\[1pt]

\textbf{\SProp} & \SPropShort & Changes to instruction that require additional knowledge about physical properties of a task-relevant object.
& referencing objects based on color, mass, size \\
\graymidrule
\textbf{\SRephrase} & \SRephraseShort & Simple rephrasing of the instruction that does not affect underlying behavior.
& verb synonyms, removing articles (e.g., ``pick up the carrot" $\to$ ``pick up carrot") \\
\graymidrule
\textbf{\SMulti} & \SMultiShort & Changes to instruction that involve references to spatial relationships between multiple objects without changing behavior.
& understanding ``left", ``right", ``in" an object (e.g., ``pick up carrot" $\to$ ``pick up object in sink") \\
\graymidrule
\textbf{\SAff} & \SAffShort & Changes to instruction that require knowledge of human affordances, or how humans interact with an object. 
& understanding human comfort, object use cases (e.g., ``hand me something I can use to clean up this mess") \\
\graymidrule
\textbf{\SInternet} & \SInternetShort & Changes to instruction that require external knowledge that can be found on the internet, and do not fall under the other \textgbf{semantic} axes.
& famous nouns (e.g., celebrities), common knowledge (e.g., tennis balls are green), typos \\

\rowcolor{RoyalBlue!20}\specialrule{1pt}{2pt}{2pt}
\multicolumn{4}{c}{\small Behavioral} \\[1pt]

\textbf{\BUnObj} & \BUnObjShort & Unobserved changes to task-relevant objects that affect behavior. & task-relevant object mass, friction, fragility \\

\graymidrule
\textbf{\BUnScene} & \BUnSceneShort & Unobserved changes to scene elements that affect behavior. & surface friction, temperature \\

\rowcolor{BlueGreen!20} \specialrule{1pt}{2pt}{2pt}
\multicolumn{4}{c}{\small Visual + Behavioral} \\[1pt]

\textbf{\VBPose} & \VBPoseShort & Changes to task-relevant object poses in the scene. & manipulated object pose, other object pose \\
\graymidrule
\textbf{\VBScene} & \VBSceneShort & Changes to scene elements that affect behavior. & clutter, surface height \\
\graymidrule
\textbf{\VBObject} & \VBObjectShort & Changes to task-relevant objects that affect their geometry. & manipulated object size, shape \\
\graymidrule
\textbf{\VBRobot} & \VBRobotShort & Changes to the robot embodiment that affect behavior. & new robot arm, new gripper or hand\\
\graymidrule
\textbf{\VBSym} & \VBSymShort & Specific to bimanual embodiments, symmetry captures changes that require the robot to mirror behavior across arms. & using different arm to perform same absolute motion, flipped absolute motion \\

\rowcolor{Orchid!20} \specialrule{1pt}{2pt}{2pt}
\multicolumn{4}{c}{\small Semantic + Behavioral} \\[1pt]

\textbf{\SBAdv} & \SBAdvShort & Changes to instruction motion descriptors that affect behavior. & speed (e.g., ``quickly" or ``slowly") \\

\graymidrule
\textbf{\SBClause} & \SBClauseShort & Changes to instruction that involve references to spatial relationships between multiple objects, which change the task specification to involve new behavior. & changing goal location for object (e.g., ``put carrot on plate" $\to$ ``put carrot in sink")  \\

\graymidrule
\textbf{\SBGrounding} & \SBGroundingShort & Changes to task-relevant nouns in the instruction to other nouns already present in the scene. & changing manipulated object to another in scene (e.g., "pick carrot" $\to$ "pick knife" when both in scene) \\

\graymidrule
\textbf{\SBAction} & \SBActionShort & Changes to action verbs in the instruction that require new behavior. & new action on a manipulated object (e.g., "pick bottle" $\to$ "rotate bottle") \\

\rowcolor{Tan!50} \specialrule{1pt}{2pt}{2pt}
\multicolumn{4}{c}{\small Visual + Semantic} \\[1pt]

\textbf{\VSObj} & \VSObjShort & Changes to task-relevant object properties that affect object appearance and language instruction, but not behavior. & new object color when base language instruction refers to the object color \\

\rowcolor{Gray!20} \specialrule{1pt}{2pt}{2pt}
\multicolumn{4}{c}{\small Visual + Semantic + Behavioral} \\[1pt]

\textbf{\VSBObj} & \VSBObjShort & Changes to task-relevant objects to new objects with different appearances, semantic descriptions, and physical characteristics. & new manipulated object (e.g., carrot $\to$ zucchini) \\

\bottomrule

\end{tabular}

\vspace{-12.5pt}
\end{table*}

\smallskip \noindent \textbf{Perturbations.} We define a perturbation function as a transformation $P: T \to T$, that applies a task delta to a base task $\tau$ to produce a new task $\tau_P$. We categorize perturbations induced by a perturbation function based on how the inputs and outputs of a policy $\pi(a \mid o^n, l)$ are impacted:

\begin{itemize}[leftmargin=*]
    \item \textbf{Visual}: $\tau_P$ is a visual perturbation of $\tau$ if $p_{\tau}(o_0) \neq p_{\tau_P}(o_0)$ (the initial distribution of image observations has changed.) %
    \item \textbf{Semantic}: $\tau_P$ is a semantic perturbation of $\tau$ if $l_\tau \neq l_{\tau_P}$ (the language instruction has changed.) %
    \item \textbf{Behavioral}: $\tau_P$ is a behavioral perturbation of $\tau$ if the expert policy $\pi_E$ changes its action distribution for the task (the required optimal behavior changes.) %
\end{itemize}

\noindent This categorization is not mutually exclusive, meaning a perturbation can fall under more than one category. %
This can also be extended to other policy modalities, e.g., if a policy uses tactile information or sound~\cite{Lee2018MakingSO}, we can further categorize perturbations based on changes to these modalities.

\subsection{Additional Assumptions}

\smallskip \noindent \textbf{Atomic Perturbations.} We focus on \emph{atomic} perturbations, which we loosely define as involving a single change (e.g., ``pick up plate" $\to$ ``push the cup" is not \emph{atomic} because it involves both changing ``plate" to ``cup" and ``pick" to ``push"). %

\smallskip \noindent \textbf{Short-Horizon Tasks.} There are forms of generalization specific to long-horizon manipulation, such as reordering sub-tasks in a sequence. We do not consider this, and instead focus on perturbations that are broadly applicable to short-horizon tasks. However, in our case study on bimanual manipulation (\cref{sec:aloha}), we evaluate on longer-horizon tasks to demonstrate that \TaxonomyName can also be applied to such settings.

\section{Axes of Generalization}
\setlength{\tabcolsep}{0.62pt}
\begin{table*}[!ht]
\vspace{5pt}
\centering
\fontsize{7}{8}\selectfont

\renewcommand{\arraystretch}{1.2}
\begin{tabular}{p{0.25cm}p{1.76cm}*{17}{>{\centering\arraybackslash}p{0.65cm}}}
\toprule
& &  \multicolumn{4}{c}{\cellcolor{OliveGreen!20}\small Visual} & \multicolumn{5}{c}{\cellcolor{Apricot!40}\small Semantic} & \multicolumn{1}{c}{\cellcolor{RoyalBlue!20}\small B} & \multicolumn{4}{c}{\cellcolor{BlueGreen!20}\small VB} & \multicolumn{2}{c}{\cellcolor{Orchid!20}\small SB} & \multicolumn{1}{c}{\cellcolor{Gray!20}\small VSB}\\
 & & \cellcolor{OliveGreen!20}\tiny\textbf{\AugShorter} & \cellcolor{OliveGreen!20}\tiny\textbf{\VSceneShorter} & \cellcolor{OliveGreen!20}\tiny\textbf{\VObjectShorter} & \cellcolor{OliveGreen!20}\tiny\textbf{\ViewShorter} & \cellcolor{Apricot!40}\tiny\textbf{\SPropShorter} & \cellcolor{Apricot!40}\tiny\textbf{\SRephraseShorter} & \cellcolor{Apricot!40}\tiny\textbf{\SMultiShorter} & \cellcolor{Apricot!40}\tiny\textbf{\SAffShorter} & \cellcolor{Apricot!40}\tiny\textbf{\SInternetShorter} & \cellcolor{RoyalBlue!20}\tiny\textbf{\BUnObjShorter} & \cellcolor{BlueGreen!20}\tiny\textbf{\VBPoseShorter} & \cellcolor{BlueGreen!20}\tiny\textbf{\VBSceneShorter} & \cellcolor{BlueGreen!20}\tiny\textbf{\VBObjectShorter} & \cellcolor{BlueGreen!20}\tiny\textbf{\VBRobotShorter} & \cellcolor{Orchid!20}\tiny\textbf{\SBClauseShorter} & \cellcolor{Orchid!20}\tiny\textbf{\SBGroundingShorter} & \cellcolor{Gray!20}\tiny\textbf{\VSBObjShorter} \\
\midrule

\multirow{7}{*}{\rotatebox[origin=c]{90}{\scriptsize\textbf{Simulation Data}}} 
& FactorWorld~\cite{xie2024decomposing}           &\fcheck &\fcheck &\fcheck &\fcheck &        &        &        &        &        &        &\fcheck &\fcheck &\fcheck &        &        &        &        \\
& KitchenShift~\cite{xing2021kitchenshift}          &\fcheck &        &\fcheck &\fcheck &        &        &        &        &        &        &\fcheck &        &\fcheck &\fcheck &        &        &        \\
& Colosseum~\cite{pumacay2024colosseum}             &\fcheck &\fcheck &\fcheck &\fcheck &        &        &        &        &        &\fcheck &        &        &\fcheck &        &        &        &        \\
& Eff-Comp~\cite{@gao2024} &        &\fcheck & \fcheck  &\fcheck &        &        &        &        &        &        &\fcheck &        &        &        &        &        &        \\
& MimicLabs~\cite{saxena2025what}     &        &  \fcheck      &  \fcheck      &  \fcheck      &        &        &        &        &        &        & \fcheck &  & &        &        &        &        \\
& CALVIN~\cite{mees2022calvin}                &        &\fcheck &\fcheck &        &\fcheck &\fcheck &        &        &        &        &\fcheck &\fcheck &\fcheck &        &        &        &        \\
& VLABench~\cite{zhang2024vlabench}              &        &        &        &        &\fcheck &\fcheck &\fcheck &\fcheck &\fcheck &        &        &        &        &        &\fcheck &        &\fcheck \\
\graymidrule

\multirow{3}{*}{\rotatebox[origin=c]{90}{\scriptsize\textbf{Real Data}}} 
& Scaling~\cite{lin2024data}     &        &        &\fcheck &        &        &        &        &        &        &        &\fcheck &\fcheck &\fcheck &        &        &        &        \\
& BridgeV2~\cite{walke2023bridgedata}     &\fcheck &\fcheck &\fcheck &        &        &        &        &        &        &\fcheck &\fcheck &\fcheck &\fcheck &\fcheck &        &        &\fcheck \\
& DROID~\cite{khazatsky2024droid}                 &        &\fcheck &\fcheck &\fcheck &        &        &        &        &        &        &        &        &\fcheck &\fcheck &        &        &        \\
\midrule

\multirow{5}{*}{\rotatebox[origin=c]{90}{\scriptsize\textbf{Policy}}} 
& BC-Z~\cite{jang2022bc}                  &        &\fcheck &        &        &        &        &        &        &        &        &\fcheck &        &\fcheck &\fcheck &        &        &\fcheck \\
& RT-Series~\cite{brohan2022rt,brohan2023rt}             &        &\fcheck &        &        &\fcheck &        &\fcheck &        &\fcheck &        &\fcheck &\fcheck &\fcheck &\fcheck &\fcheck &        &\fcheck \\
& MT-ACT~\cite{bharadhwaj2023roboagent}                &\fcheck &\fcheck &        &        &        &        &        &        &        &        &\fcheck &\fcheck &\fcheck &\fcheck &        &        &\fcheck \\
& $\pi_0$~\cite{black2024pi_0}               &        &\fcheck &\fcheck &        &        &        &        &        &        &        &\fcheck &\fcheck &\fcheck &\fcheck &        &        &\fcheck \\
& OpenVLA~\cite{kim24openvla}               &        &\fcheck &        &        &\fcheck &\fcheck &\fcheck &        &\fcheck &        &\fcheck &        &\fcheck &\fcheck &        &\fcheck &\fcheck \\

\midrule
& \BenchName         &        &\fcheck &\fcheck &\fcheck &\fcheck &\fcheck &\fcheck &        &\fcheck &        &\fcheck &\fcheck &\fcheck &        &\fcheck &\fcheck &\fcheck \\
\bottomrule

\end{tabular}
\caption{\small We present existing generalization benchmarks/datasets and generalist policy learning works through the lens of \TaxonomyName, including \BenchName. Columns are different axes in \TaxonomyName (a subset of 17/22 axes in \cref{tab:axes}). %
A checkmark indicates a given axis is considered.} %
\label{tab:compare}
\vspace{-10pt}
\end{table*} 

\subsection{\TaxonomyName: A Taxonomy of Generalization}
\label{sec:axes:tax}
Here we define \TaxonomyName, our taxonomy of generalization. We aim to organize perturbations in a human-interpretable manner to guide policy evaluation. To this end, we define \emph{factors}, \emph{axes}, and \emph{categories}, which represent different levels of hierarchy in our taxonomy, in decreasing order of granularity.

\smallskip \noindent \textbf{Factors.} We define a factor as a human-interpretable, fine-grained grouping of perturbations that affect a task in a common way. For example, if the lighting in a scene is changed in multiple ways, each would represent a separate perturbation. We can then group all such perturbations under the factor ``Lighting". Factors can be categorized as \textgbf{visual}, \textgbf{semantic}, and/or \textgbf{behavioral}, based on their constituent perturbations. %

\smallskip \noindent \textbf{Axes.} We define an axis as a human-interpretable grouping of similar factors that affect a common set of policy modalities. For example, our taxonomy defines \emph{Image Augmentations} as an axis of \textgbf{visual} factors that can be varied using simple image transforms, such as ``Lighting" or ``Image Blur". The axes in our taxonomy are designed to be a practically comprehensive set of the most salient challenges identified in the literature. %

\smallskip \noindent \textbf{Categories.} We define a category as a grouping of all axes that affect the same combination of policy modalities. For example, the category \textgbf{visual} captures all axes that only affect the initial image observations of a task, including \emph{Image Augmentations}. There are seven possible categories (the number of combinations of policy modalities). By capturing all combinations of how policy modalities can be affected by a given perturbation, we intend for these categories to provide a complete framing of generalization conditions for robot manipulation.

\smallskip We outline the axes in \TaxonomyName in \cref{tab:axes}. For each axis, we provide a description and examples of constituent factors. While these axes are intended to be applicable to a broad range of tasks, some tasks will not always have meaningful instantiations of an axis. In \cref{fig:axes}, we show example perturbations of the base task ``put carrot on plate" for several axes. %

\subsection{Prior Notions of Generalization}
\label{sec:axes:comparison}

In \cref{tab:compare}, we list prior works that measure generalization and the axes of \TaxonomyName they consider, in comparison with \BenchName, a benchmark we designed using \TaxonomyName (\cref{sec:measuring}). As shown, \TaxonomyName aims to be a comprehensive superset of prior efforts to measure generalization. There are other nuances between prior works and \TaxonomyName that are observable in \cref{tab:compare}, some of which we describe here.

Prior works are often not as fine-grained as \TaxonomyName in their categorization of generalization. For example, RT-2~\cite{brohan2023rt} simply groups many of our \textgbf{visual + behavior} axes under the blanket category ``Behavior Generalization". Prior works also often categorize perturbations in ways that are only applicable to certain tasks. For example, Colosseum~\cite{pumacay2024colosseum} considers ``Receiver Object" (RO) perturbations (e.g., the ``rack" in ``put wine in rack"), which does not apply to tasks without such objects. In \TaxonomyName, we address this by considering different levels of hierarchy, where our high-level categorization is amenable to all tasks, while our lower levels may be more task-specific. %

Lastly, prior works differ in how their evaluation protocols consider perturbations, often in less practical ways. For example, OpenVLA~\cite{kim24openvla} considers perturbations with respect to a portion of training data from a large-scale mixture (OXE), for a scene that their evaluation setting aims to emulate. However, it can be difficult to replicate scenes from an outside data source, possibly leading to unintended perturbations. In \BenchName, we define \TaxonomyName perturbations with respect to in-domain data from the evaluation scene, to more easily control for this.

\label{sec:axes}

\section{Case Study 1: Bridge V2}
\label{sec:measuring}
\TaxonomyName provides a broad framework for generalization in robot manipulation, but how should it be used to evaluate policies? In this section, we use \TaxonomyName to instantiate \BenchName, a real-world benchmark based on Bridge V2~\cite{walke2023bridgedata}. We first describe our benchmark and our rationale for its design. Then, we use it to evaluate several state-of-the-art open-source models and variations. Our goal is for \BenchName to demonstrate how \TaxonomyName can be used to design generalization benchmarks using a reproducible and open-source platform.

\subsection{Instantiating \texorpdfstring{\TaxonomyName}{} on Bridge V2}
\label{sec:measuring:instantiation}

\smallskip \noindent \textbf{Dataset.} We use Bridge V2~\cite{walke2023bridgedata} as our pre-training dataset and platform, since it has been used in multiple prior works to study generalization~\cite{walke2023bridgedata, kim24openvla, team2024octo}, and its training environments have been reliably reproduced for evaluation~\cite{yang2024robot, hejna2024re, team2024octo, kim24openvla}.

\smallskip \noindent \textbf{Base Tasks.} We consider the base tasks ``put carrot on plate", ``put knife on plate", ``flip pot upright", and ``put plate in sink".
We choose these base tasks based on the support of the pre-training data. In particular, they are instantiated in a replication of a sink environment from Bridge V2 that was used to evaluate generalization in prior work~\cite{kim24openvla}. %
We choose these specific tasks to cover different levels of alignment with the original tasks from Bridge V2 for this sink environment. %

\begin{figure}[!h]
\setlength{\tabcolsep}{1pt}
\renewcommand{\arraystretch}{1.1}
\fontsize{8}{8}\selectfont
\centering
\begin{tabular}{p{0.3cm}ccc}
  \toprule
  & Axis & Factors \\
 \midrule
\multirow{2}{*}{\rotatebox[origin=r]{90}{\scriptsize\textgbf{Visual}}} 
 & \VSceneShort & distractors, surface color \\
 & \VObjectShort & other object color \\
 & \ViewShort & camera pose \\
 \graymidrule
\multirow{4}{*}{\rotatebox[origin=c]{90}{\scriptsize\textgbf{Semantic}}} 
 & \SPropShort & referencing color \\
 & \SRephraseShort & changing verbs \\
 & \SMultiShort & understanding ``in" and ``on" \\
 & \SInternetShort & common obj properties, typos \\
 \graymidrule
\multirow{3}{*}{\rotatebox[origin=c]{90}{\scriptsize\textgbf{VB}}} 
 & \VBPoseShort & manipulated object pose \\
 & \VBSceneShort & surface height \\
 & \VBObjectShort & manipulated object size, shape \\
 \graymidrule
\multirow{2}{*}{\rotatebox[origin=c]{90}{\scriptsize\textgbf{SB}}} 
 & \SBClauseShort & understanding ``in" \\
 & \SBActionShort & new action on object \\
 \graymidrule
\multirow{1}{*}{\rotatebox[origin=c]{90}{\scriptsize\textgbf{VSB}}}
 & \VSBObjShort & new manipulated object \\[4pt]
 \bottomrule
\end{tabular}
\caption{\small Axes and factors in \BenchName.}
\vspace{-7.5pt}
\label{tab:bridge_axes} 
\end{figure}

\begin{figure}[t!]
    \centering
    \vspace{2pt}
    \includegraphics[width=0.9\columnwidth]{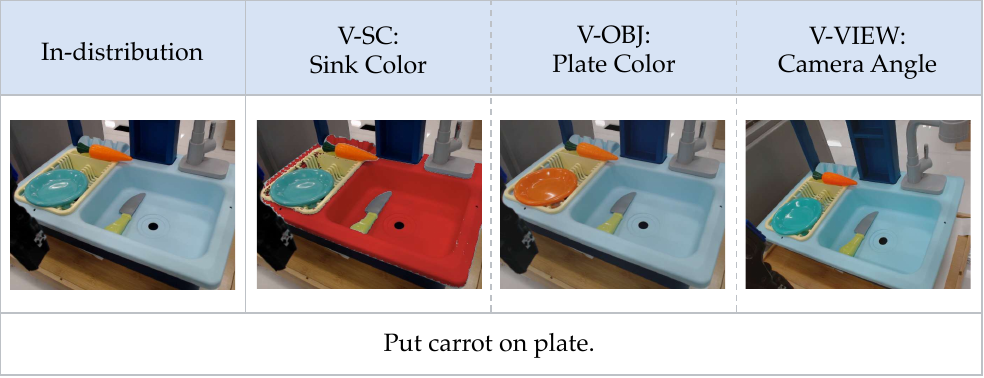}
    \caption{\small Examples of \textgbf{visual} perturbations in \BenchName. Left: in-distribution base task scene. From left to right: we vary sink color (V-SC), plate color (V-OBJ), and camera angle (V-VIEW).}
    \label{fig:visual_bridge}
    \vspace{-5pt}
\end{figure}

\begin{figure}[t!]
    \centering
    \includegraphics[width=0.6\columnwidth]{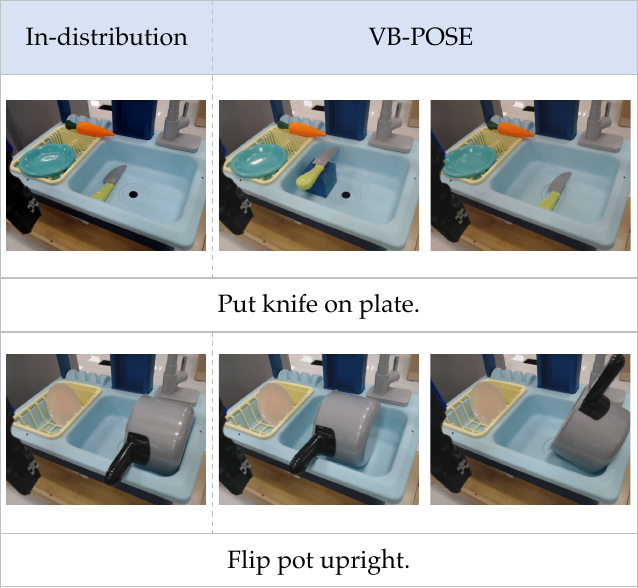}
    \caption{\small Examples of object poses (\VBPoseShort) in \BenchName.}
    \label{fig:visual_behavioral_bridge}
    \vspace{-10pt}
\end{figure}

\smallskip \noindent \textbf{Evaluation Conditions.}
We evaluate 4 in-distribution base tasks and 55 perturbations that cover 13/22 axes in \TaxonomyName. We do not cover some axes due to incompatibility with our base tasks. We list our evaluated axes and factors in \cref{tab:bridge_axes}, and visualize some base tasks and their perturbations in \cref{fig:visual_bridge} and \cref{fig:visual_behavioral_bridge}. We further detail our evaluation conditions in our Appendix (can be found on our \href{stargen-taxonomy.github.io}{website}). %

\smallskip \noindent \textbf{Policies.} We focus our evaluation on state-of-the-art open-source imitation learning policies that have demonstrated generalization for Bridge V2 tasks in prior work. Specifically, we analyze three vision-language-action (VLA) models that fine-tune foundation models on robot data: OpenVLA \cite{kim24openvla}, MiniVLA \cite{belkhale2024minivla}, and a third-party reimplementation of $\pi_0$ \cite{black2024pi_0, ren2024pi}. These models cover a range of design decisions that reflect the state of generalist manipulation policies.

\smallskip \noindent \textbf{Evaluation Procedure.} We evaluate our models using a co-fine-tuning procedure. First, we pre-train each model only on Bridge V2. Next, we collect base task demonstrations from our evaluation environment, and co-fine-tune on this with Bridge V2. We denote co-fine-tuned models with (FT).

For the ``put carrot" and ``put knife" base tasks, we collect 10 demonstrations per base task. For the ``flip pot" and ``put plate" base tasks, we collect 50 demonstrations per base task. We execute policies until the robot succeeds, reaches a dangerous/irrecoverable state, or terminates after 100 timesteps. We perform five trials per condition for each model. %

\begin{figure}[!h]
    \centering
    \includegraphics[width=0.8\columnwidth]{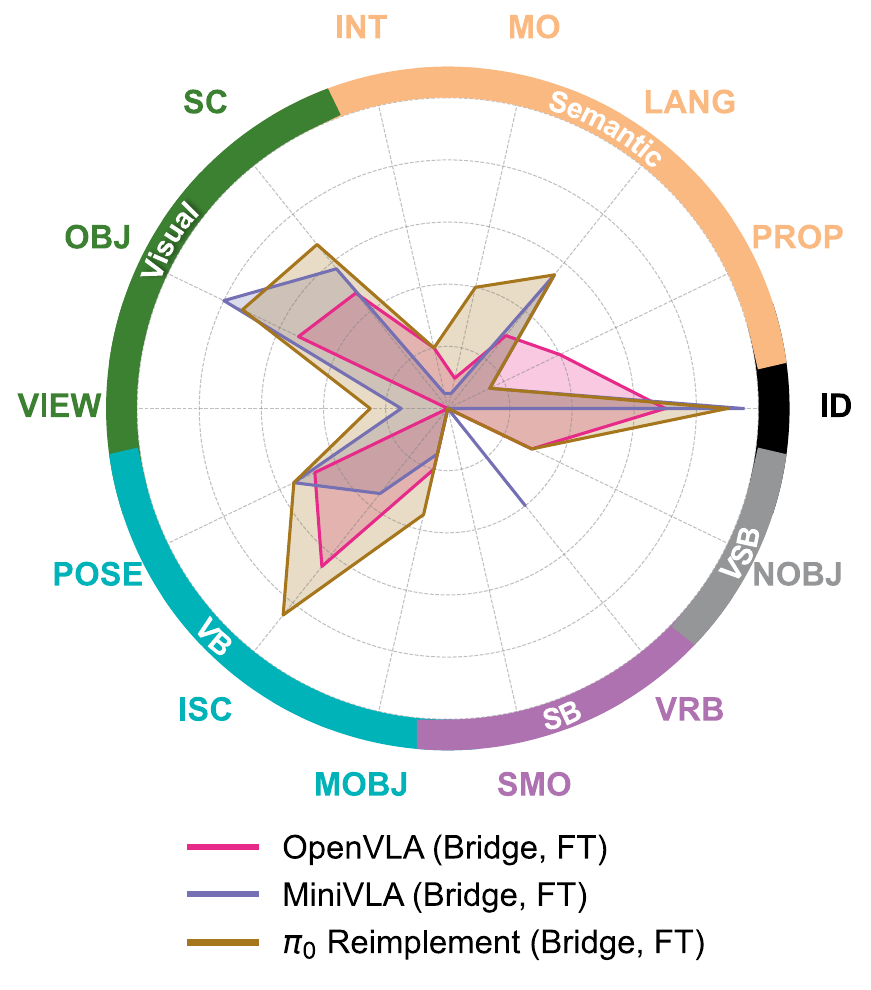}
    \caption{\small \BenchName main results. We report aggregated success rates for each model and axis, including in-distribution (ID).}
    \label{fig:main-results}
    \vspace{-15pt}
\end{figure}

\begin{figure*}[!ht]
    \centering
    \vspace{2pt}
    \begin{subfigure}[t]{0.24\textwidth}
        \includegraphics[width=\textwidth]{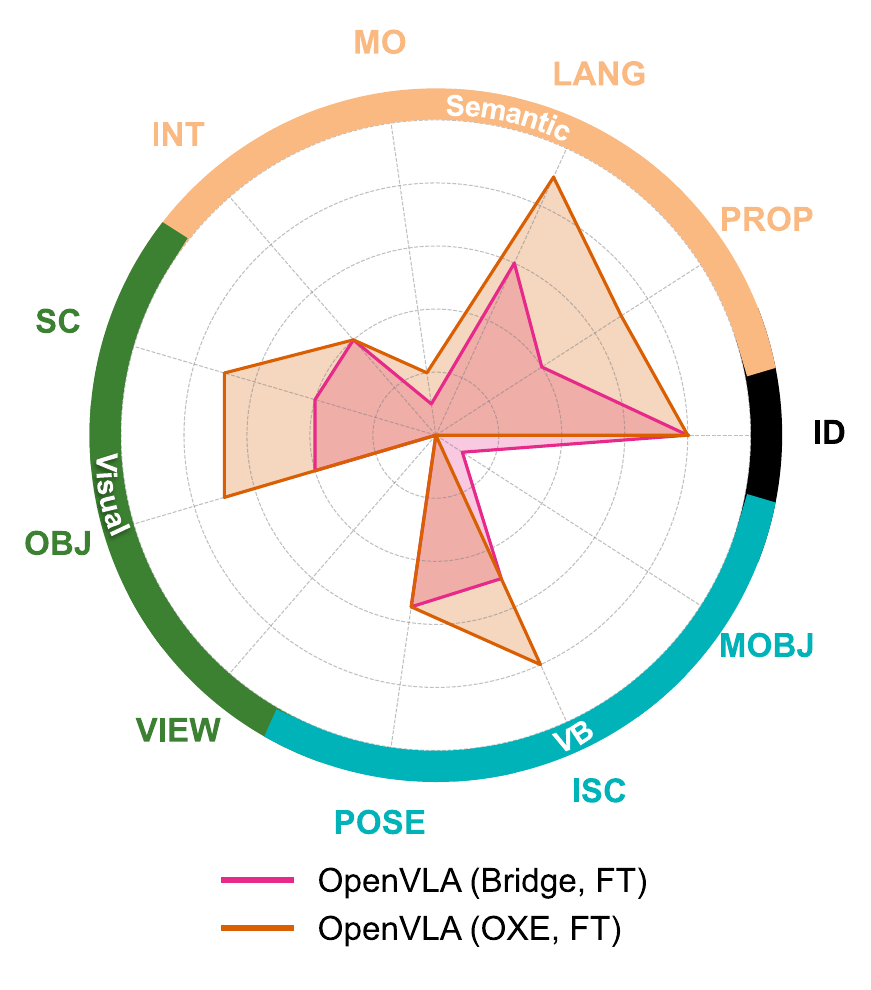}
        \vspace{-20pt}
        \caption{\small Scaling robot data}
    \end{subfigure}
    \begin{subfigure}[t]{0.24\textwidth}
        \includegraphics[width=\textwidth]{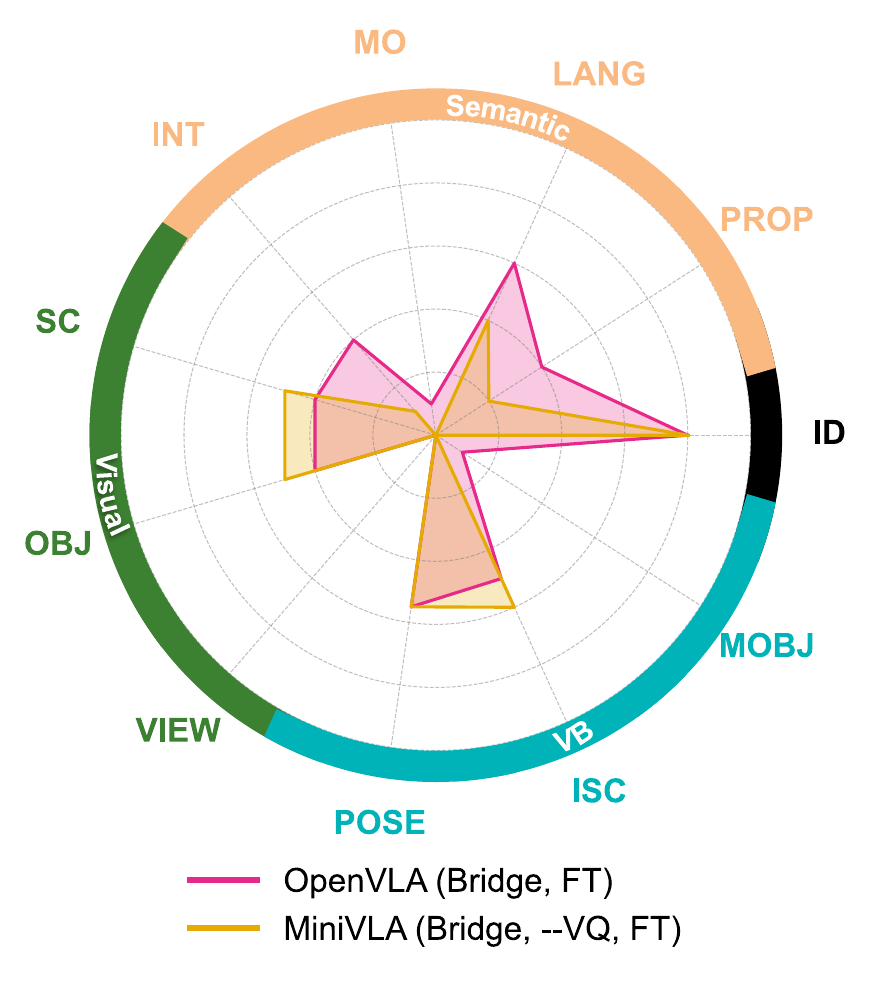}
        \vspace{-20pt}
        \caption{\small Scaling LLM backbones}
    \end{subfigure}
    \begin{subfigure}[t]{0.24\textwidth}
        \includegraphics[width=\textwidth]{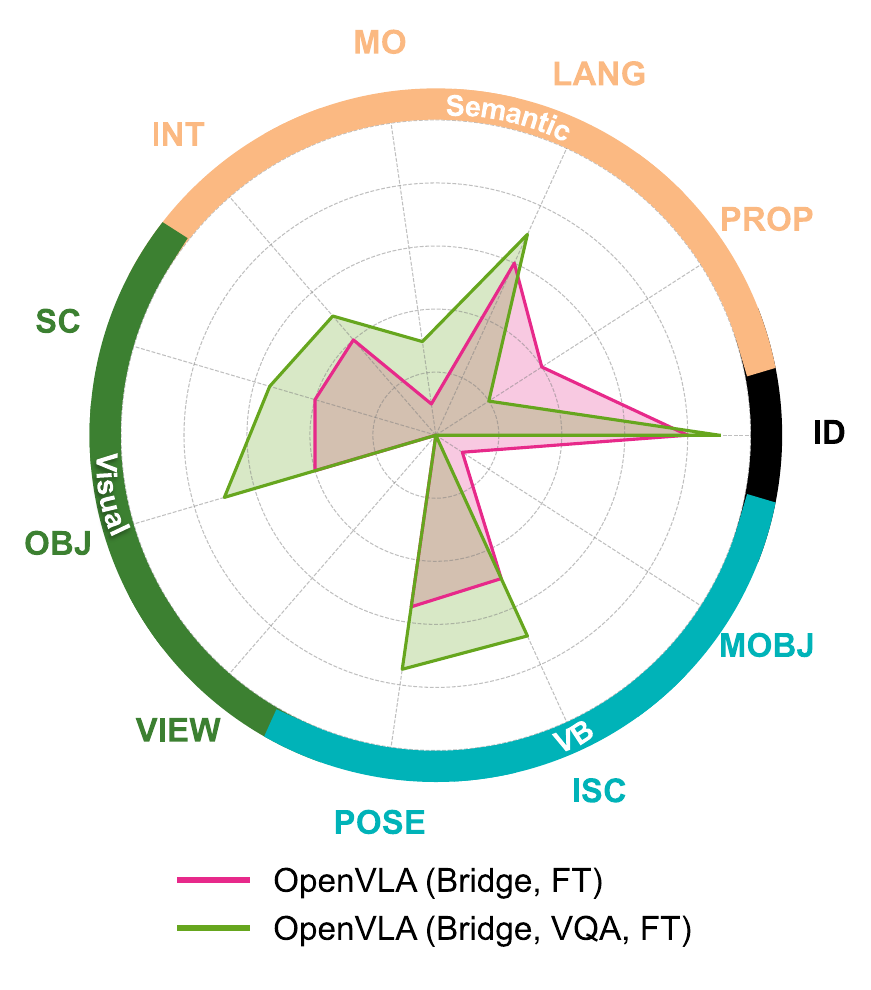}
        \vspace{-20pt}        
        \caption{\small VQA co-training}
    \end{subfigure}
    \begin{subfigure}[t]{0.24\textwidth}
        \includegraphics[width=\textwidth]{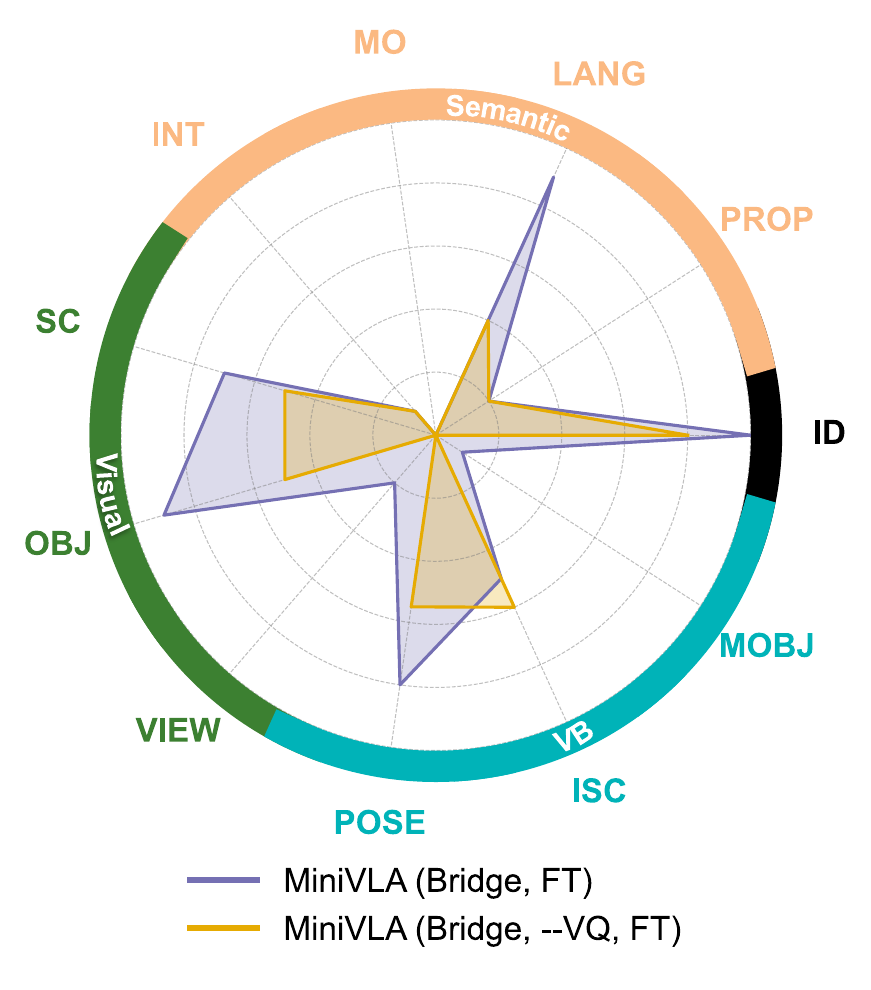}
        \vspace{-20pt}
        \caption{\small VQ action chunking}
    \end{subfigure}
    \caption{\small We investigate VLA design decisions. (a) Scaling robot datasets can help. (b) Larger LLMs can provide a modest benefit to \textgbf{semantic} axes. (c) VQA co-training can help, but has a mixed effect on \textgbf{semantic} axes. (d) VQ action chunking can help.}
    \label{fig:ablations-1}
    \vspace{-10pt}
\end{figure*}

\subsection{Main Results}
\label{sec:measuring:results}

In \cref{fig:main-results}, we report our main results on \BenchName, which consist of 885 trials. We find that existing generalist policies tend to struggle on most axes. In particular, \textgbf{semantic} generalization is mostly weak, despite the use of language model backbones. %
This has interesting implications: e.g., instead of relying only on language model initialization to improve \textgbf{semantic} generalization, perhaps other mechanisms are needed, such as improving robot language annotations~\cite{smith2024steer}. %

Each model tends to have similar strengths and weaknesses. However, there are some notable differences that the fine-grained nature of our benchmark helps reveal. For example, OpenVLA is noticeably worse at \textgbf{visual} generalization, while MiniVLA struggles more with \textgbf{visual + behavioral}. OpenVLA is the best at understanding object properties, %
but still struggles with other \textgbf{semantic} axes. $\pi_0$ generally performs the best, possibly due to a more capable VLM backbone (PaliGemma \cite{beyer2024paligemma}), and/or better architecture (flow-based action chunking). %

\begin{table}[h]
    \centering
    \begin{tabular}{c@{\hskip 5pt}c@{\hskip 5pt}c@{\hskip 5pt}c@{\hskip 5pt}c}
        \toprule
         Visual & Semantic & \makecell{Visual +\\Behavioral} & \makecell{Semantic +\\Behavioral} & \makecell{Across\\Categories} \\
        \midrule
         0.84 & -0.07 & 0.36 & 0.87 & 0.02 \\
        \bottomrule
    \end{tabular}
    \caption{\small Average Pearson correlations of performance for axes within the same category (left) and across categories.  (right)}
    \label{tab:axes_correlations}
    \vspace{-5pt}
\end{table}

\smallskip \noindent \textbf{Axes Correlations.} To further motivate our high-level categorization based on policy modalities, we investigate performance correlations across models for axes within the same category, compared to across categories. In \cref{tab:axes_correlations}, we find that correlations are higher within categories, except for \textgbf{semantic}. We hypothesize this is because our \textgbf{semantic} axes can require much different forms of reasoning (e.g., understanding object properties is much different than language rephrasing).

\smallskip \noindent \textbf{Prioritizing Axes.} From these results, we provide some general guidelines on axes to prioritize in future work.

\begin{itemize}
    \item \textgbf{Visual}-only axes generally exhibit stronger generalization than \textgbf{behavioral} axes (with the exception of \emph{Viewpoint}). We hypothesize this is because VLA vision-language pre-training is more likely to convey visual robustness than generalization to new behavior. Therefore, future work on generalist manipulation should de-prioritize \textgbf{visual}-only axes (except \emph{Viewpoint}) in favor of 
    \textgbf{behavioral} axes.
    \item While \textgbf{semantic} generalization is weak, whether to prioritize these axes depends on how the policy is deployed. If the policy is used with open-ended language, then these axes are important. However, if language is more constrained (e.g., a system where a separate model provides a limited set of instructions), they can be de-prioritized.
\end{itemize}

\subsection{Investigating VLA Design Decisions}
\label{sec:measuring:ablations}
While our main results provide insights on model capabilities, it is difficult to disentangle what contributes to generalization. To better understand this, we conduct additional targeted evaluations on model design choices, with $t$-tests to assess statistical significance. %

\smallskip \noindent \textbf{Scaling Robot Data.} In \cref{fig:ablations-1}(a) we compare our Bridge-only OpenVLA with a version trained on a significantly larger, cross-embodiment OXE mixture~\cite{o2023open}. Consistent with prior work~\cite{o2023open, kim24openvla}, we find that larger and more diverse datasets can significantly improve forms of generalization, such as for \textgbf{visual + behavioral} axes $(M=0.22$ vs. $M=0.48), t(7) = -2.76, p = 0.028$. However, the axes on which the Bridge-only model struggled the most (\emph{\View}, \emph{\VBObject}, \emph{\SMulti}) do not improve significantly.

\smallskip \noindent \textbf{Scaling LLM Backbones.} In \cref{fig:ablations-1}(b) we compare VLA policies that differ only in the large language model (LLM) backbone. Specifically, we compare OpenVLA (Bridge, FT), using Llama 2 7B ~\cite{touvron2023llama}, and MiniVLA (Bridge, --VQ, FT), using Qwen2.5 0.5B~\cite{yang2024qwen2}. The only major difference between these two models is their LLM backbone. We find that while the larger LLM improves \textgbf{semantic} axes, it is not by a significant amount $(M=0.18$ vs. $M=0.35), t(7) = -1.87, p = 0.104$. Absolute performance for these and other axes also remain low, suggesting that scaling LLMs only has limited benefits.

\smallskip \noindent \textbf{VQA Co-training.} In \cref{fig:ablations-1}(c), we investigate co-training with visual-question answering (VQA) data, which prior work has shown to improve generalization~\cite{brohan2023rt}. We find this can help, such as for \textgbf{visual} axes $(M=0.30$ vs. $M=0.45), t(7) = -2.39, p = 0.048$. However, there is surprisingly a mixed effect for \textgbf{semantic} axes $(M=0.38$ vs. $M=0.42), t(7) = -0.51, p = 0.626$, improving three of them, but hurting \emph{\SProp}. This indicates room for improvement, possibly by using data targeted for embodied reasoning~\cite{zawalski2024robotic}. 

\smallskip \noindent \textbf{VQ Action Chunking}. In \cref{fig:ablations-1}(d), we investigate using binning-based tokenization instead of vector quantized action chunking with MiniVLA. We find that VQ action chunking helps nearly all axes, including \textgbf{visual} axes by a significant amount $(M=0.38$ vs. $M=0.62), t(7) = -2.38, p = 0.049$. This highlights the importance of action chunking and tokenization methods, as also suggested by prior work~\cite{zhao2023learning, pertsch2025fast}.

\section{Case Study 2: Bimanual Manipulation}
\label{sec:aloha}
Next, we use \TaxonomyName to develop an additional case study based on the bimanual ALOHA 2 platform~\cite{aldaco2024aloha}.

\begin{figure}[ht!]
    \centering
    \includegraphics[width=0.65\columnwidth]{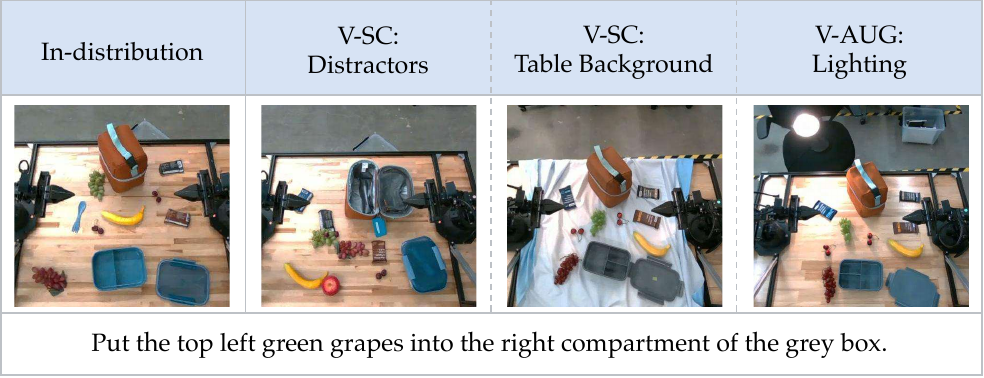}
    \caption{\small Examples of \textgbf{visual} perturbations in our bimanual case study. Left: in-distribution base task. From left to right: we vary distractor objects (V-SC), table background (V-SC), and lighting (V-AUG).}
    \label{fig:visual}
    \vspace{-15pt}
\end{figure}

\subsection{Experimental Setup}
\begin{figure}[t!]
    \centering
    \vspace{2pt}
    \includegraphics[width=0.9\columnwidth]{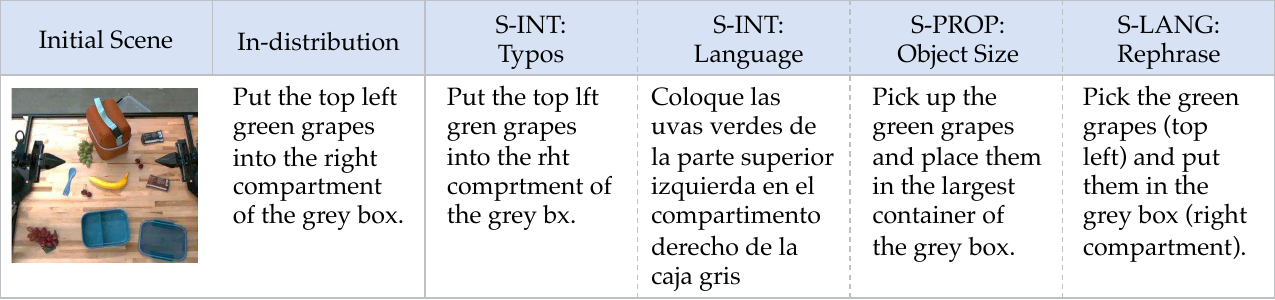}
    \caption{\small Examples of \textgbf{semantic} perturbations in our bimanual case study. Left: in-distribution base task instruction. From left to right: we test robustness to typos (S-INT), language (S-INT), understanding object size (S-PROP), and rephrasing (S-LANG).}
    \vspace{-3pt}
    \label{fig:semantic}
\end{figure}

\begin{figure}[t!]
    \centering
    \includegraphics[width=0.9\columnwidth]{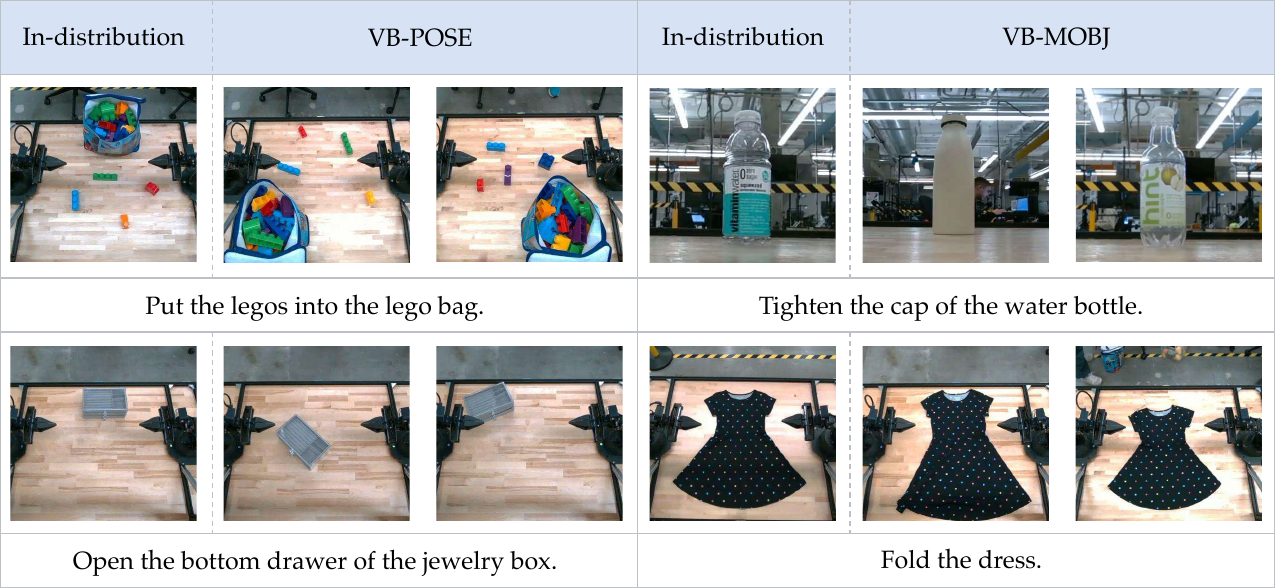}
    \caption{\small Examples of \textgbf{visual + behavioral} perturbations in our bimanual case study. Left: changes to object pose (VB-POSE) from in-distribution. Right: changes to object geometry (VB-MOBJ).}
    \vspace{-3pt}
    \label{fig:visual_behavioral}
\end{figure}

We use a proprietary robot dataset of thousands of hours of teleoperated demonstrations collected on a fleet of ALOHA 2 robots. This allows us to assess generalization for tasks with more variety, dexterity, and horizon length than those considered in \BenchName, such as tightening a water bottle and folding a dress. We evaluate on 17 in-distribution base tasks with 68 perturbations that cover 7 axes in \TaxonomyName. These are the same conditions used to evaluate generalization for Gemini Robotics~\cite{team2025gemini} (see Appendix C.1.3 in the technical report), but recategorized according to \TaxonomyName axes. We visualize examples of conditions for \textgbf{visual} axes in \cref{fig:visual}, \textgbf{semantic} axes in \cref{fig:semantic}, and \textgbf{visual + behavioral} axes in \cref{fig:visual_behavioral}.

We evaluate 3 models: a multi-task diffusion policy~\cite{chi2023diffusion}, a reimplementation of $\pi_0$~\cite{black2024pi_0}, and Gemini Robotics On-Device (GRoD)~\cite{deepmind2025gemini_robotics_on_device}, a proprietary VLA. We report policy task progress. %
We refer to~\cite{team2025gemini} for more details on our evaluation conditions, protocol, and the diffusion and $\pi_0$ policies.

\subsection{Results}
\begin{figure}[!ht]
    \centering
    \includegraphics[width=0.8\columnwidth]{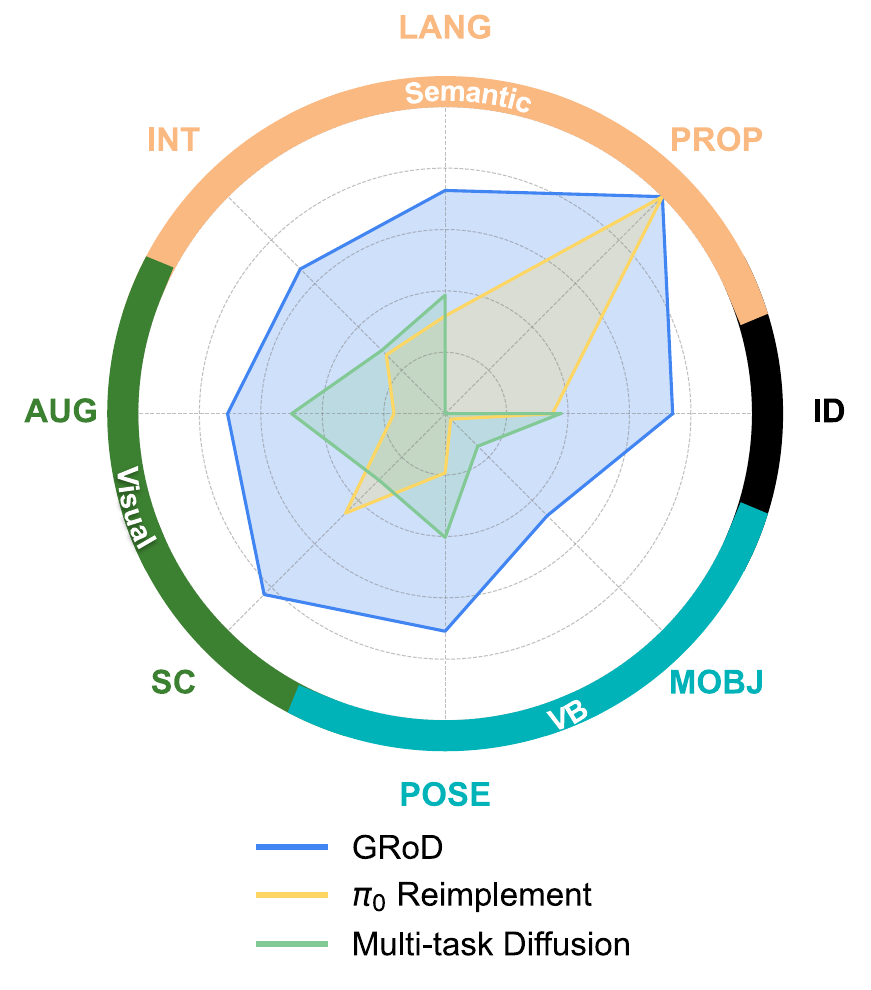}
    \vspace{-3pt}
    \caption{\small Results from our bimanual manipulation case study. We report aggregated task progress for each model and axis.}
\label{fig:aloha_results}
    \vspace{-15pt}
\end{figure}

We report our results in \cref{fig:aloha_results}, which consist of 390 trials. We find that GRoD consistently outperforms the other models, which we speculate is due to its strong VLM backbone and other architectural enhancements. In particular, \TaxonomyName helps identify perturbations that GRoD is robust to while the other models fail almost entirely, such as translating the instruction to a new language (example in \cref{fig:semantic}). We use this case study to demonstrate the broad applicability of \TaxonomyName with more diverse, dexterous, and long-horizon tasks.

\section{Discussion}
\label{sec:discussion}
We present \TaxonomyName, a taxonomy of generalization for robot manipulation. Our taxonomy not only thoroughly considers the space of visuo-lingual policy generalization, but is also straightforward to instantiate. We demonstrate the considerations and design process for instantiating \TaxonomyName on a real-world, reproducible benchmark%
, eliciting key insights about generalist policy capabilities design choices. We further demonstrate using \TaxonomyName to study generalization for a wider set of tasks and policies on %
the ALOHA 2 platform. We hope that \TaxonomyName can help improve the comprehensiveness of generalization benchmark design to generate better insights.

\smallskip \noindent \textbf{Limitations and Future Work.} Due to constraints imposed by real-world evaluation time (1600+ trials), we only evaluate a subset of axes and factors that we believe most effectively demonstrate the utility of \TaxonomyName. We hope that \TaxonomyName can guide future benchmarking efforts that expand the scope considered in our evaluations, such as using simulation to more efficiently and comprehensively measure policy generalization.

While we consider \TaxonomyName to be a strong starting point, we believe that future work can revise and expand our taxonomy based on the needs of robotics practitioners. Specifically, while we have argued for the completeness of our high-level categories, our set of fine-grained axes is empirically derived. Future work may identify new generalization challenges that could be incorporated as new axes within our framework.

\bibliographystyle{IEEEtran}
\bibliography{references}

@string{iclr = {International Conference on Learning Representations (ICLR)}}

@string{icml = {International Conference on Machine Learning (ICML)}}

@string{icra = {International Conference on Robotics and Automation (ICRA)}}

@string{rss = {Proceedings of Robotics: Science and Systems (RSS)}}

@string{neurips = {Advances in Neural Information Processing Systems (NeurIPS)}}

@string{corl = {Conference on Robot Learning (CoRL)}}

@string{cvpr = {Conference on Computer Vision and Pattern Recognition (CVPR)}}

@string{ijrr = {The International Journal of Robotics Research (IJRR)}}

@inproceedings{walke2023bridgedata,
  title={{BridgeData V2: A Dataset for Robot Learning at Scale}},
  author={Walke, Homer Rich and Black, Kevin and Zhao, Tony Z and Vuong, Quan and Zheng, Chongyi and Hansen-Estruch, Philippe and He, Andre Wang and others},
  booktitle={Conference on Robot Learning (CoRL)},
  year={2023},
}

@inproceedings{dasari2019robonet,
    title={{RoboNet: Large-Scale Multi-Robot Learning}},
    author={Sudeep Dasari and Frederik Ebert and Stephen Tian and Suraj Nair and Bernadette Bucher and Karl Schmeckpeper and Siddharth Singh and Sergey Levine and Chelsea Finn},
    year={2019},
    booktitle={Conference on Robot Learning (CoRL)},
}

@inproceedings{jang2022bc,
  title={{BC-Z: Zero-Shot Task Generalization with Robotic Imitation Learning}},
  author={Jang, Eric and Irpan, Alex and Khansari, Mohi and Kappler, Daniel and Ebert, Frederik and Lynch, Corey and Levine, Sergey and others},
  booktitle={Conference on Robot Learning (CoRL)},
  year={2022},
}

@inproceedings{ebert2021bridge,
  title={{Bridge Data: Boosting Generalization of Robotic Skills with Cross-Domain Datasets}},
  author={Ebert, Frederik and Yang, Yanlai and Schmeckpeper, Karl and Bucher, Bernadette and Georgakis, Georgios and Daniilidis, Kostas and Finn, Chelsea and Levine, Sergey},
  booktitle={Proceedings of Robotics: Science and Systems (RSS)},
  year={2022}
}

@inproceedings{fang2023rh20t,
  author={Fang, Hao-Shu and Fang, Hongjie and Tang, Zhenyu and Liu, Jirong and others},
  booktitle={International Conference on Robotics and Automation (ICRA)}, 
  title={{RH20T: A Comprehensive Robotic {D}ataset for Learning Diverse Skills in One-Shot}}, 
  year={2024},
}

@inproceedings{khazatsky2024droid,
  title={{DROID: A Large-Scale In-the-Wild Robot Manipulation Dataset}},
  author={Khazatsky, Alexander and Pertsch, Karl and Nair, Suraj and Balakrishna, Ashwin and Dasari, Sudeep and Karamcheti, Siddharth and others},
  booktitle={Proceedings of Robotics: Science and Systems (RSS)},
  year={2024}
}

@inproceedings{o2023open,
  title={{Open X-Embodiment: Robotic Learning Datasets and RT-X Models}},
  author={OXE Collaboration and others},
  booktitle={International Conference on Robotics and Automation (ICRA)},
  year={2024},
}

@inproceedings{kim24openvla,
    title = 	 {{OpenVLA: An Open-Source Vision-Language-Action Model}},
    author={{Moo Jin} Kim and Karl Pertsch and Siddharth Karamcheti and Ted Xiao and Ashwin Balakrishna and Suraj Nair and Rafael Rafailov and others},
    booktitle={Conference on Robot Learning (CoRL)},
    year={2024},
}

@inproceedings{bharadhwaj2023roboagent,
    title={{RoboAgent: Generalization and Efficiency in Robot Manipulation via Semantic Augmentations and Action Chunking}},
    author={Homanga Bharadhwaj and Jay Vakil and Mohit Sharma and Abhinav Gupta and Shubham Tulsiani and Vikash Kumar},
    booktitle={International Conference on Robotics and Automation (ICRA)},
    year={2024},
}

@inproceedings{team2024octo,
  title={{Octo: An Open-Source Generalist Robot Policy}},
  author={Team, Octo Model and Ghosh, Dibya and Walke, Homer and Pertsch, Karl and Black, Kevin and Mees, Oier and Dasari, Sudeep and Hejna, Joey and others},
  booktitle={Proceedings of Robotics: Science and Systems (RSS)},
  year={2024}
}

@inproceedings{brohan2022rt,
  title={{RT-1: Robotics Transformer for Real-World Control at Scale}},
  author={Brohan, Anthony and Brown, Noah and Carbajal, Justice and Chebotar, Yevgen and Dabis, Joseph and Finn, Chelsea and others},
  booktitle={Proceedings of Robotics: Science and Systems (RSS)},
  year={2023}
}

@inproceedings{brohan2023rt,
  title={{RT-2: Vision-Language-Action Models Transfer Web Knowledge to Robotic Control}},
  author={Brohan, Anthony and Brown, Noah and Carbajal, Justice and Chebotar, Yevgen and Chen, Xi and Choromanski, Krzysztof and others},
  booktitle = 	 {Conference on Robot Learning},
  year = 	 {2023},
}

@inproceedings{
    saxena2025what,
    title={{What Matters in Learning from Large-Scale Datasets for Robot Manipulation}},
    author={Vaibhav Saxena and Matthew Bronars and Nadun Ranawaka Arachchige and Kuancheng Wang and others},
    booktitle={International Conference on Learning Representations (ICLR)},
    year={2025},
}

@inproceedings{lin2024data,
  title={{Data Scaling Laws in Imitation Learning for Robotic Manipulation}},
  author={Lin, Fanqi and Hu, Yingdong and Sheng, Pingyue and Wen, Chuan and You, Jiacheng and Gao, Yang},
  booktitle={International Conference on Learning Representations (ICLR)},
  year={2025},
}

@inproceedings{@gao2024,
  title={{Efficient Data Collection for Robotic Manipulation via Compositional Generalization}},
  author={Jensen Gao and Annie Xie and Ted Xiao and Chelsea Finn and Dorsa Sadigh},
  booktitle={Proceedings of Robotics: Science and Systems (RSS)},
  year={2024}
}

@inproceedings{pumacay2024colosseum,
  title={{THE COLOSSEUM: A Benchmark for Evaluating Generalization for Robotic Manipulation}},
  author={Pumacay, Wilbert and Singh, Ishika and Duan, Jiafei and Krishna, Ranjay and others},
  booktitle={Proceedings of Robotics: Science and Systems (RSS)},
  year={2024}
}

@inproceedings{smith2024steer,
      title={{STEER: Flexible Robotic Manipulation via Dense Language Grounding}}, 
      author={Laura Smith and Alex Irpan and Montserrat Gonzalez Arenas and Sean Kirmani and Dmitry Kalashnikov and Dhruv Shah and Ted Xiao},
        booktitle={International Conference on Robotics and Automation (ICRA)},
      year={2025},
}

@inproceedings{Lee2018MakingSO,
  title={{Making Sense of Vision and Touch: Self-Supervised Learning of Multimodal Representations for Contact-Rich Tasks}},
  author={Michelle A. Lee and others},
  booktitle={International Conference on Robotics and Automation (ICRA)},
  year={2018},
}

@inproceedings{xie2024decomposing,
  title={{Decomposing the Generalization Gap in Imitation Learning for Visual Robotic Manipulation}},
  author={Xie, Annie and Lee, Lisa and Xiao, Ted and Finn, Chelsea},
  booktitle={IEEE International Conference on Robotics and Automation (ICRA)},
  year={2024},
}

@inproceedings{xing2021kitchenshift,
  title={{KitchenShift: Evaluating Zero-Shot Generalization of Imitation-Based Policy Learning Under Domain Shifts}},
  author={Xing, Eliot and Gupta, Abhinav and Powers, Sam and Dean, Victoria},
  booktitle={NeurIPS 2021 Workshop on Distribution Shifts: Connecting Methods and Applications},
  year={2021}
}

@inproceedings{zhao2023learning,
  title={{Learning Fine-Grained Bimanual Manipulation with Low-Cost Hardware}},
  author={Zhao, Tony Z and Kumar, Vikash and Levine, Sergey and Finn, Chelsea},
  booktitle={Proceedings of Robotics: Science and Systems (RSS)},
  year={2023}
}

@inproceedings{li2024evaluating,
    title={{Evaluating Real-World Robot Manipulation Policies in Simulation}},
    author={Li, Xuanlin and Hsu, Kyle and Gu, Jiayuan and Pertsch, Karl and Mees, Oier and Walke, Homer Rich and Fu, Chuyuan and Lunawat, Ishikaa and Sieh, Isabel and others},
    booktitle={Conference on Robot Learning (CoRL)},
    year={2024},
}

@misc{belkhale2024minivla,
      title={{MiniVLA: A Better VLA with a Smaller Footprint}}, 
      author={Suneel Belkhale and Dorsa Sadigh},
      url={https://github.com/Stanford-ILIAD/openvla-mini},
      year={2024},
}

@misc{ren2024pi,
      title={{Open Pi-Zero}}, 
      author={Allen Ren},
      url={https://github.com/allenzren/open-pi-zero},
      year={2024},
}

@inproceedings{liu2024improved,
  title={{Improved Baselines with Visual Instruction Tuning}},
  author={Liu, Haotian and Li, Chunyuan and Li, Yuheng and Lee, Yong Jae},
  booktitle={Conference on Computer Vision and Pattern Recognition (CVPR)},
  year={2024}
}

@inproceedings{karamcheti2024prismatic,
  title = {{Prismatic VLMs: Investigating the Design Space of Visually-Conditioned Language Models}},
  author = {Siddharth Karamcheti and Suraj Nair and Ashwin Balakrishna and others},
  booktitle = {International Conference on Machine Learning (ICML)},
  year = {2024},
}

@article{ravi2024sam2,
  title={{SAM 2: Segment Anything in Images and Videos}},
  author={Ravi, Nikhila and Gabeur, Valentin and Hu, Yuan-Ting and Hu, Ronghang and Ryali, Chaitanya and Ma, Tengyu and Khedr, Haitham and R{\"a}dle, Roman and Rolland, Chloe and Gustafson, Laura and Mintun, Eric and Pan, Junting and others},
  journal={arXiv preprint arXiv:2408.00714},
  year={2024}
}

@inproceedings{mandlekar2018roboturk,
    title={{RoboTurk: A Crowdsourcing Platform for Robotic Skill Learning through Imitation}},
    author={Mandlekar, Ajay and Zhu, Yuke and Garg, Animesh and Booher, Jonathan and Spero, Max and Tung, Albert and Gao, Julian and others},
    booktitle={Conference on Robot Learning (CoRL)},
    year={2018},
}

@inproceedings{mirchandani2024robocrowd,
  title={{RoboCrowd: Scaling Robot Data Collection through Crowdsourcing}},
  author={Mirchandani, Suvir and Yuan, David D and Burns, Kaylee and Islam, Md Sazzad and Zhao, Tony Z and others},
  booktitle={International Conference on Robotics and Automation (ICRA)},
  year={2025}
}

@inproceedings{black2024pi_0,
  title={{$\pi_0$: A Vision-Language-Action Flow Model for General Robot Control}},
  author={Black, Kevin and Brown, Noah and Driess, Danny and Esmail, Adnan and Equi, Michael and Finn, Chelsea and Fusai, Niccolo and others},
  booktitle={Proceedings of Robotics: Science and Systems (RSS)},
  year={2025}
}

@article{mu2021maniskill,
  title = {{ManiSkill: Generalizable Manipulation Skill Benchmark with Large-Scale Demonstrations}},
  author={Mu, Tongzhou and Ling, Zhan and Xiang, Fanbo and Yang, Derek and Li, Xuanlin and Tao, Stone and Huang, Zhiao and Jia, Zhiwei and Su, Hao},
  booktitle = {Proceedings of the Neural Information Processing Systems Track on Datasets and Benchmarks},
  year={2021},
}

@article{mees2022calvin,
  title={{CALVIN: A Benchmark for Language-Conditioned Policy Learning for Long-Horizon Robot Manipulation Tasks}},
  author={Mees, Oier and Hermann, Lukas and others},
  journal={IEEE Robotics and Automation Letters (RA-L)},
  year={2022},
}

@inproceedings{nasiriany2024robocasa,
  title={{RoboCasa: Large-Scale Simulation of Everyday Tasks for Generalist Robots}},
  author={Nasiriany, Soroush and Maddukuri, Abhiram and Zhang, Lance and Parikh, Adeet and Lo, Aaron and Joshi, Abhishek and others},
  booktitle={Proceedings of Robotics: Science and Systems (RSS)},
  year={2024}
}

@inproceedings{szot2021habitat,
  title={{Habitat 2.0: Training Home Assistants to Rearrange their Habitat}},
  author={Szot, Andrew and Clegg, Alexander and Undersander, Eric and Wijmans, Erik and Zhao, Yili and Turner, John and others},
  booktitle={Advances in Neural Information Processing Systems (NeurIPS)},
  year={2021}
}

@inproceedings{li2023behavior,
  title={{BEHAVIOR-1K: A Benchmark for Embodied AI with 1,000 Everyday Activities and Realistic Simulation}},
  author={Li, Chengshu and Zhang, Ruohan and Wong, Josiah and Gokmen, Cem and Srivastava, Sanjana and others},
  booktitle={Conference on Robot Learning (CoRL)},
  year={2023},
}

@inproceedings{ehsani2021manipulathor,
  title={{ManipulaTHOR: A Framework for Visual Object Manipulation}},
  author={Ehsani, Kiana and Han, Winson and Herrasti, Alvaro and VanderBilt, Eli and Weihs, Luca and Kolve, Eric and others},
  booktitle={Conference on Computer Vision and Pattern Recognition (CVPR)},
  year={2021}
}

@article{james2020rlbench,
  title={{RLBench: The Robot Learning Benchmark \& Learning Environment}},
  author={James, Stephen and Ma, Zicong and Arrojo, David Rovick and Davison, Andrew J},
  journal={IEEE Robotics and Automation Letters (RA-L)},
  year={2020},
}

@article{zhang2024vlabench,
  title={{VLABench: A Large-Scale Benchmark for Language-Conditioned Robotics Manipulation with Long-Horizon Reasoning Tasks}},
  author={Zhang, Shiduo and Xu, Zhe and Liu, Peiju and Yu, Xiaopeng and others},
  journal={arXiv},
  year={2024}
}

@inproceedings{hejna2024re,
    title={{Re-Mix: Optimizing Data Mixtures for Large Scale Imitation Learning}},
    author={Hejna, Joey and Bhateja, Chethan and Jian, Yichen and Pertsch, Karl and Sadigh, Dorsa},
    booktitle={Conference on Robot Learning (CoRL)},
    year={2024},
}

@inproceedings{yang2024robot,
  title={{Robot Fine-Tuning Made Easy: Pre-Training Rewards and Policies for Autonomous Real-World Reinforcement Learning}},
  author={Yang, Jingyun and others},
  booktitle={International Conference on Robotics and Automation (ICRA)},
  year={2024},
}

@article{touvron2023llama,
  title={{Llama 2: Open Foundation and Fine-Tuned Chat Models}},
  author={Touvron, Hugo and Martin, Louis and Stone, Kevin and Albert, Peter and Almahairi, Amjad and Babaei, Yasmine and Bashlykov, Nikolay and Batra, Soumya and Bhargava, Prajjwal and others},
  journal={arXiv},
  year={2023}
}

@article{yang2024qwen2,
  title={{Qwen2.5 Technical Report}},
  author={Yang, An and Yang, Baosong and Zhang, Beichen and Hui, Binyuan and Zheng, Bo and Yu, Bowen and Li, Chengyuan and Liu, Dayiheng and Huang, Fei and Wei, Haoran and others},
  journal={arXiv},
  year={2024}
}

@inproceedings{wang2024scaling,
    author    = {Lirui Wang and Xinlei Chen and Jialiang Zhao and Kaiming He},
    title     = {{Scaling Proprioceptive-Visual Learning with Heterogeneous Pre-trained Transformers}},
    booktitle = {Advances in Neural Information Processing Systems (NeurIPS)},
    year      = {2024}
}

@inproceedings{
    liu2025rdtb,
    title={{RDT-1B: a Diffusion Foundation Model for Bimanual Manipulation}},
    author={Songming Liu and Lingxuan Wu and Bangguo Li and Hengkai Tan and Huayu Chen and Zhengyi Wang and Ke Xu and Hang Su and others},
    booktitle={International Conference on Learning Representations (ICLR)},
    year={2025},
}

@article{shafiullah2023bringing,
  title={{On Bringing Robots Home}},
  author={Shafiullah, Nur Muhammad Mahi and Rai, Anant and Etukuru, Haritheja and Liu, Yiqian and Misra, Ishan and Chintala, Soumith and Pinto, Lerrel},
  journal={arXiv},
  year={2023}
}

@article{kress2024robot,
  title={{Robot Learning as an Empirical Science: Best Practices for Policy Evaluation}},
  author={Kress-Gazit, Hadas and Hashimoto, Kunimatsu and Kuppuswamy, Naveen and Shah, Paarth and Horgan, Phoebe and others},
  journal={arXiv preprint arXiv:2409.09491},
  year={2024}
}

@article{muller2007dynamic,
  title={{Dynamic Time Warping}},
  author={M{\"u}ller, Meinard},
  journal={Information retrieval for music and motion},
  pages={69--84},
  year={2007},
  publisher={Springer}
}

@article{beyer2024paligemma,
  title={{PaliGemma: A versatile 3B VLM for transfer}},
  author={Beyer, Lucas and Steiner, Andreas and Pinto, Andr{\'e} Susano and Kolesnikov, Alexander and Wang, Xiao and Salz, Daniel and Neumann, Maxim and Alabdulmohsin, Ibrahim and Tschannen, Michael and others},
  journal={arXiv},
  year={2024}
}

@article{aldaco2024aloha,
  title={{ALOHA 2: An Enhanced Low-Cost Hardware for Bimanual Teleoperation}},
  author={Aldaco, Jorge and Armstrong, Travis and Baruch, Robert and Bingham, Jeff and Chan, Sanky and Draper, Kenneth and Dwibedi, Debidatta and others},
  journal={arXiv},
  year={2024}
}

@inproceedings{pertsch2025fast,
  title={{FAST: Efficient Action Tokenization for Vision-Language-Action Models}},
  author={Pertsch, Karl and Stachowicz, Kyle and Ichter, Brian and Driess, Danny and Nair, Suraj and Vuong, Quan and others},
  booktitle={Proceedings of Robotics: Science and Systems (RSS)},
  year={2025}
}

@article{chi2023diffusion,
  title={{Diffusion Policy: Visuomotor Policy Learning via Action Diffusion}},
  author={Chi, Cheng and Xu, Zhenjia and Feng, Siyuan and Cousineau, Eric and Du, Yilun and Burchfiel, Benjamin and others},
  journal={The International Journal of Robotics Research (IJRR)},
  year={2023},
}

@inproceedings{zawalski2024robotic,
  title={{Robotic Control via Embodied Chain-of-Thought Reasoning}},
  author={Zawalski, Micha{\l} and Chen, William and Pertsch, Karl and Mees, Oier and Finn, Chelsea and Levine, Sergey},
  booktitle={Conference on Robot Learning (CoRL)},
  year={2024}
}

@misc{deepmind2025gemini_robotics_on_device,
  author       = {{Google DeepMind}},
  title        = {{Gemini Robotics On-Device}},
  year         = {2025},
  url = {https://deepmind.google/models/gemini-robotics/gemini-robotics-on-device/},
}

@article{team2025gemini,
  title={{Gemini Robotics: Bringing AI into the Physical World}},
  author={Team, Gemini Robotics and Abeyruwan, Saminda and Ainslie, Joshua and Alayrac, Jean-Baptiste and Arenas, Montserrat Gonzalez and Armstrong, Travis and Balakrishna, Ashwin and Baruch, Robert and Bauza, Maria and Blokzijl, Michiel and others},
  journal={arXiv preprint arXiv:2503.20020},
  year={2025}
}

\clearpage
\appendices
\label{sec:appendix}
\section{Taxonomy Categorization}
\label[appendix]{sec:axes_descriptions}
Here, we provide more detailed descriptions of how we categorize the axes and factors in \TaxonomyName based on the combinations of policy modalities they affect.

\smallskip \noindent \textbf{Visual}: Factors and axes in this category modulate the image inputs of the initial state, but do not affect the required behavior in the base task or the language instruction. Example factors include lighting, camera pose, and distractor objects (see the green sector of \cref{fig:axes}).

\smallskip \noindent \textbf{Semantic}: Semantic factors and axes modulate the language instruction without changing the initial image or required behavior. Example factors include replacing verbs with synonyms, and changing the instruction to use spatial relationships such as ``in the \rule{0.5cm}{0.15mm}" (see the orange sector of \cref{fig:axes}).

\smallskip \noindent \textbf{Behavioral}: Behavioral factors and axes only affect required behavior without affecting policy inputs. Therefore, all isolated behavioral factors are necessarily \emph{unobserved} from single observations. Example factors include changes to object mass or friction (see the blue sector of \cref{fig:axes}). These factors are often challenging for policies due to their unobservability.

\smallskip \noindent \textbf{Visual + Behavioral}: These factors and axes affect the initial image and required behavior, without changing the language instruction. As we show in \cref{sec:axes:comparison}, many factors that prior work consider as ``behavior" generalization fall into this category~\cite{kim24openvla,brohan2023rt}. Example factors include manipulated object poses and surface/table heights (see the cyan sector of \cref{fig:axes}).

\smallskip \noindent \textbf{Semantic + Behavioral}: These factors and axes affect the language instruction and required behavior, without affecting the initial image. Example factors include changing the speed of a behavior (``quickly" vs. ``slowly") in language, or specifying prepositional phrases like ``into" or ``in front of" that change the required behavior (see the purple sector of \cref{fig:axes}).

\smallskip \noindent \textbf{Visual + Semantic}: These factors and axes affect the initial image and language instruction, without requiring change to behavior. An example of this would be if the base instruction is ``pick up the purple cup" and the cup changes color to blue, changing the instruction to ``pick up the blue cup". This is still an \emph{atomic} perturbation, but since the color was specified in the original instruction, the instruction must also be changed. Had the initial instruction been ``pick up the cup" and the cup was already blue, then this would be a \textgbf{semantic} perturbation. See the brown sector of \cref{fig:axes}.

\smallskip \noindent \textbf{Visual + Semantic + Behavioral}: This category of factors and axes affect all three modalities at once. An example is going from ``pick up the carrot" to ``pick up the zucchini" -- this affects the initial image, the language, and the behavior required to pick up the new object. %

\section{Frequently Asked Questions}
\label[appendix]{sec:faq}
Here, we provide answers to several possible questions readers may have about our taxonomy and its assumptions.
\newcommand{\Question}[1]{\medskip \noindent \emph{Q: #1}}
\newcommand{\Answer}[1]{\smallskip \noindent \emph{A}: #1}

\Question{Is \TaxonomyName meant to be comprehensive?}

\Answer{The axes presented in \TaxonomyName are meant as a starting point for the field, and although we did our best to make this list comprehensive, there could certainly exist other meaningful axes. In contrast, we consider the \emph{categories} in \TaxonomyName (e.g., \textgbf{visual + behavioral}) to be exhaustive for the policies we consider, since they are built from the seven unique combinations of our policy modalities.}

\Question{Are the axes and factors in \TaxonomyName subjective?}

\Answer{Yes, the axes and factors in \TaxonomyName are human-specified in a subjective manner, so there are certainly other ways to group perturbations into factors, and factors into axes. However, our goal is not to design an objective way to categorize generalization, as this is an inherently subjective process. Instead, we aim to provide greater structure and comprehensiveness when categorizing generalization using human interpretable concepts, like ``task-relevant objects" and ``verbs".}

\Question{What is the purpose of defining a taxonomy if it is inherently subjective?}

\Answer{As we will see in \cref{sec:measuring}, our taxonomy gives us the vernacular to discuss more fine-grained types of generalization. We intend this taxonomy to be a starting point for practitioners to gain greater insights about their models, and recognize the potential for reshaping our taxonomy as our understanding of how to effectively evaluate generalization grows.}

\Question{Is each axis equally important for generalization?}

\Answer{Whether or not an axis is ``important" is very subjective, and depends on specific applications. Instead, we categorize different types of generalization more systematically to aid researchers and practitioners in evaluating their robot policies based on the needs of their downstream applications.
}

\Question{Can you quantify how much different perturbations affect a base task?}

\Answer{It would be interesting to consider the ``edit distance" that a perturbation induces, to quantify how much generalization is required. However, we do not consider this in \TaxonomyName, due to the challenging nature of defining such distances. Some options for this could include image or text embeddings for \textgbf{visual} or \textgbf{semantic} perturbations, dynamic time warping~\cite{muller2007dynamic} for \textgbf{behavioral}, or foundation models. Future work can investigate correlating such metrics with empirical generalization.}

\section{Case Study 1: \BenchName}

\subsection{Base Tasks}
\label[appendix]{sec:base_tasks}
We consider the following four base tasks in our evaluation for \BenchName:

\begin{enumerate}
    \item Put carrot on plate.
    \item Put knife on plate.
    \item Flip pot upright.
    \item Put plate in sink.
\end{enumerate}

Each task is instantiated in a replication of a sink environment that was used in Bridge V2, and was also used to evaluate generalization in prior work~\cite{kim24openvla}. We tuned the environment conditions such that the publicly released OpenVLA 7B model trained on OXE was able to achieve reasonable zero-shot performance on the ``put carrot" and ``put knife" tasks, in order to increase the likelihood of transfer from the pre-training data for generalization. However, there are inevitably some minor differences from this environment and what was used to collect data in Bridge V2, such as in camera pose or lighting. While we could have chosen base tasks in an environment that deviates more from the pre-training data, this would have likely led to much weaker levels of generalization in our evaluation, which would make our results less informative.

The base tasks ``put carrot" and ``put knife" are both instantiated from the same initial configuration. ``flip pot" and ``put plate" are also both instantiated from the same initial configuration. We do this to increase the difficulty and realism for \textgbf{semantic} generalization, because having multiple tasks from the same initial configuration makes it so that policies must understand the language instruction to know which of the tasks to perform. Therefore, the policy cannot simply ignore the language instruction and memorize which task to perform based on the visual appearance of the scene.

We choose these specific tasks to cover different levels of alignment with the demonstrations from Bridge V2 for this sink environment. We describe how each base task deviates from the support of these demonstrations, and how the publicly released OpenVLA model performs on them, as follows:

\begin{enumerate}
    \item \textbf{Put carrot on plate}: This language instruction is found in Bridge V2 demonstrations for our sink environment. The carrot is different from the one in these demonstrations, but is of similar appearance and geometry. The plate is also different, with a different color, but similar geometry. The initial pose of the carrot and plate are significantly different than in any demonstration, but similar poses for other task-relevant objects are found in other demonstrations for this environment for similar tasks (e.g., ``put knife on cutting board"). We find that OpenVLA is able to somewhat reliably perform this task zero-shot.
    \item \textbf{Put knife on plate}: This language instruction is not found in Bridge V2 demonstrations for our sink environment, but demonstrations with similar instructions are (e.g., ``put carrot on plate", "put knife on cutting board"). The knife is different from the one in these demonstrations, but is of similar appearance and geometry. The plate is also different, with a different color, but similar geometry. Similar initial poses for task-relevant objects are found in demonstrations for the task ``put knife on cutting board" in this sink environment. We find that OpenVLA is able to somewhat reliably perform this task zero-shot.
    \item \textbf{Flip pot upright}: This language instruction is found in Bridge V2 demonstrations for our sink environment. The pot used is significantly different from the one used in these demonstrations (e.g., it is plastic instead of metal, and has different geometry). The initial pose of the pot is also somewhat different. We find that OpenVLA is able to rarely perform this task zero-shot.
    \item \textbf{Put plate in sink}: This language instruction is not found in Bridge V2 demonstrations for our sink environment. However, there is a somewhat similar language instruction in Bridge V2 for a slightly different sink environment (``put cup from anywhere into sink"). The initial pose, appearance, and geometry of the plate is significantly different than in any task demonstrations for our sink environment. We find that OpenVLA is unable to perform this task zero-shot, and exhibits largely meaningless behavior.

\end{enumerate}

We aim to minimize variation in the initial state distribution for each base task (e.g., by attempting to always initialize objects with the same pose). However, in practice there will always be some level of variation in real world experiments such as ours. We aim to minimize variation in order to simplify our experimental setup and to more easily isolate the effect of our considered perturbations. However, our taxonomy and evaluation framework is more general and can be applied to base tasks with greater variation in initial states.

\subsection{Evaluation Procedure Details}
\label[appendix]{sec:eval_details}
We perform our evaluations by executing each policy until either the policy succeeds as deemed by a human evaluator, it reaches a dangerous or irrecoverable state, or terminates after 100 timesteps. For tasks that involve placing an object on another object or surface, we consider success to be if the object is on top of the other object/surface in a stable configuration, and the object is not grasped by the robot. For tasks that involve placing an object on a plate, the plate must also not be knocked over from its initial position. For tasks that involve flipping a container upright, the base of the container must make contact with the base of the sink, the container must be in a stable configuration, and the container must not be grasped by the robot.

We aim to minimize differences between our evaluation conditions and in-domain training data, beyond the differences introduced by a specified perturbation. To do this, we periodically check the image difference between live observations and the base task demonstration data, and adjust the environment to minimize this when the difference is significant. We do this so that our evaluations are only reflective of generalization to a specified perturbation, and not other inadvertent changes~\cite{kress2024robot}.

\subsection{Co-fine-tuning vs. Fine-tuning}
For our evaluations, we incorporate in-domain data into each model by co-fine-tuning with the model's pre-training data. We do this to help preserve each model's original capabilities, to promote generalization and fair comparison of each model. However, this procedure may become impractical as robot pre-training datasets become larger in size, or if they are not openly available. One could also consider evaluating these models by only fine-tuning on in-domain data, without any pre-training data. However, this should be done carefully to avoid catestrophic forgetting. We leave exploration of best practices for fine-tuning generalist policies, and comparing the generalization of such fine-tuned policies, to future work.

\subsection{In-Domain Data}
We collect demonstrations for our in-domain data using a Meta Quest VR headset. For consistency, we collect all our data using a single experienced human teleoperator. For the ``put carrot" and ``put knife" base tasks, we collect 10 demonstrations per base task. For the ``flip pot" and ``put plate" base tasks, we collect 50 demonstrations per base task, because we found that more demonstrations were needed for satisfactory in-distribution performance of models.

\subsection{Model Details}

\noindent \textbf{General Training Details}. Unless otherwise stated, for all co-fine-tuned models (designated with FT), we co-fine-tune for 5K gradient steps (which we found was sufficient for convergence of most models) with a batch size of 256 and a learning rate of $1e-6$. We adopt a smaller learning rate than what was used during pre-training because we found higher learning rates tend to degrade training metrics on the pre-training dataset. We use the Adam optimizer for all training. When co-fine-tuning, we normalize the in-domain dataset using the dataset statistics from Bridge V2.

For each model type, we co-fine-tune two separate models: one for the ``put carrot" and ``put knife" base tasks, and another for the ``flip pot" and ``put plate" base tasks. We do this because we had already trained and partially evaluated models co-fine-tuned only on ``put carrot" and ``put knife" before expanding our base task selection to include the others. However, it would also been fine to co-fine-tune models on all base tasks together. When co-fine-tuning models for the ``put carrot" and ``put knife" base tasks, we upsample the in-domain data by 100x relative to the pre-training data. For the "flip pot" and "put plate" base tasks, we upsample the in-domain data by 50x. \\

\noindent We describe the specific training details of each model we consider in our experiments below:

\smallskip \noindent \textbf{OpenVLA (OXE)}.
We use the publicly released OpenVLA 7B model trained on a mixture of data from OXE~\cite{kim24openvla}.

\smallskip \noindent \textbf{OpenVLA (OXE, FT)}.
We co-fine-tune OpenVLA (OXE) on a combination of the same OXE mixture used to originally train the model (except DROID~\cite{khazatsky2024droid}), as well as the in-domain data for a given set of base tasks. 

\smallskip \noindent \textbf{OpenVLA (Bridge, FT)}.
We first pre-train a version of OpenVLA 7B on Bridge V2 for 200K gradient steps with a batch size of 256. We then co-fine-tune with Bridge V2.

\smallskip \noindent \textbf{OpenVLA (Bridge, Rand Init, FT)}.
We first pre-train a version of OpenVLA 7B on Bridge V2, except we initialize the model from random weights, rather than a pre-trained vision-language model. We pre-train for 200K gradient steps with a batch size of 256. We then co-fine-tune with Bridge V2.

\smallskip \noindent \textbf{OpenVLA (Bridge, VQA, FT)}.
We first pre-train a version of OpenVLA 7B on Bridge V2 and the VQA dataset used for training LLaVA-1.5 (\citep{liu2024improved}). This is a subset of the dataset used for training the base Prism-7B used for initializing OpenVLA~\cite{karamcheti2024prismatic}. We set of 20\% of our pre-training mixture to consist of VQA data. We pre-train for 200K gradient steps with a batch size of 256. We then co-fine-tune with Bridge V2, again using 20\% VQA data.

\smallskip \noindent \textbf{MiniVLA (Bridge, FT)}. We start from the publicly released MiniVLA model trained on Bridge V2~\cite{belkhale2024minivla}, and then co-fine-tune this model with Bridge V2. We increase the amount of gradient steps during co-fine-tuning from 5K to 10K, which was necessary for convergence. During inference, we only execute the first action from each predicted action chunk.

\smallskip \noindent \textbf{MiniVLA (Bridge, No VQ, FT)}.
We first pre-train a version of MiniVLA that uses the binning-based tokenizer from OpenVLA rather than the vector quantized action chunking tokenizer used in MiniVLA. We pre-train on Bridge V2 for 300K gradient steps with a batch size of 256 and learning rate of $2e-5$, and then co-fine-tune with Bridge V2.

\smallskip \noindent \textbf{$\boldsymbol{\pi_0}$ (Bridge, FT)}.
We use the $\pi_0$ reimplementation from \url{https://github.com/allenzren/open-pi-zero}~\cite{ren2024pi}. We start from the publicly released checkpoint pre-trained on Bridge V2 using beta timestep sampling, and then co-fine-tune this model with Bridge V2. We co-fine-tune for 80K training steps and a learning rate of $1e-5$, with a linear warmup of 200 steps. We found it was necessary to increase the in-domain upsampling rate to 1000x relative to Bridge V2 for ``put carrot" and ``put knife", and 500x for "flip pot" and "put plate", in order to fit the in-domain data more efficiently. During inference, we only execute the first action from each predicted action chunk.

Unlike all other models, this model was pre-trained using image augmentation (random crop, lighting, and color). During co-fine-tuning, we remove image augmentation to increase training speed. Furthermore, this model only uses trajectories from Bridge V2 that have language annotations.

\begin{figure}[!ht]
    \centering
    \includegraphics[width=0.4\textwidth]{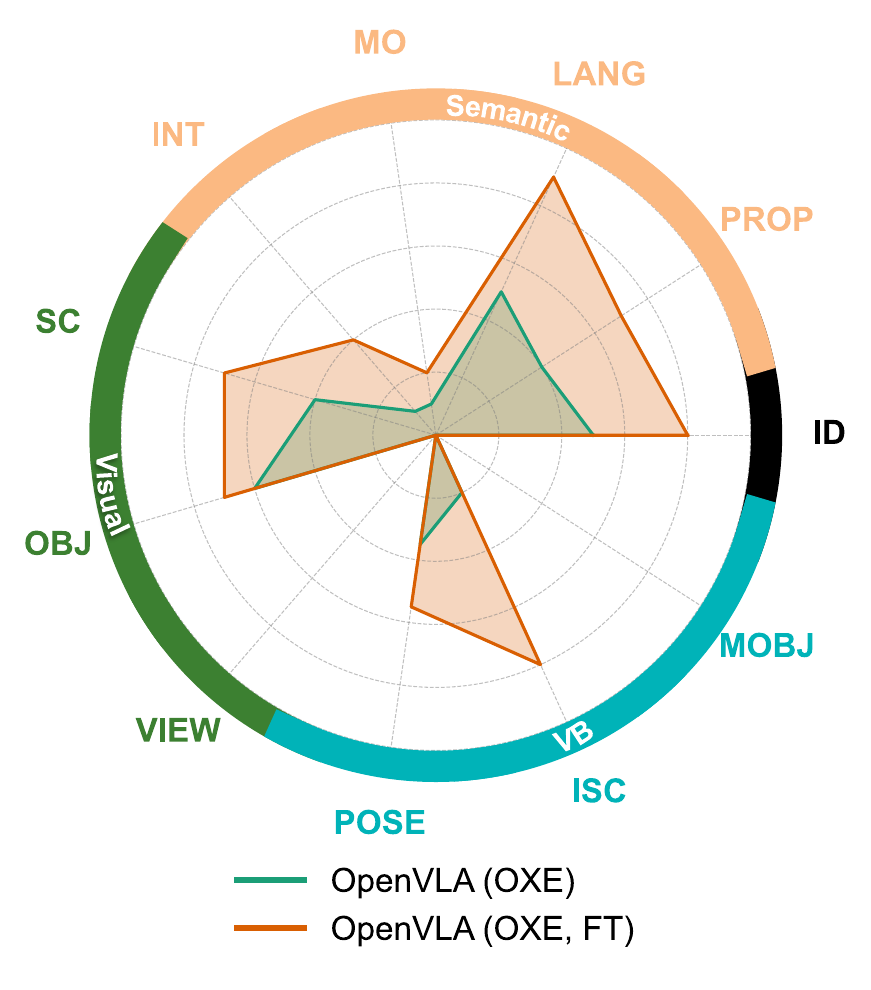}
    \caption{We find that co-fine-tuning on in-domain data for our base tasks significantly improves both in-distribution performance of OpenVLA, as well as many of our axes.}
    \vspace{-10pt}
    \label{fig:zero_shot}
\end{figure}

\subsection{Impact of In-Domain Data}
\label[appendix]{sec:zero_shot_evals}
To motivate our choice to assess generalization by comparing models co-fine-tuned on in-domain base task data, in \cref{fig:zero_shot} we compare the publicly released OpenVLA 7B model zero-shot, and the same model co-fine-tuned with in-domain data, for the ``put carrot" and ``put knife" tasks. We fine that co-fine-tuning significantly improves both in-distribution base task performance and multiple axes of generalization. This highlights the importance of co-fine-tuning on base task data in order to properly define base tasks as in-distribution. Otherwise, it is likely there will be some significant distribution shift between the pre-training data and the chosen base tasks for evaluation. As a result, performance in general will suffer, making evaluation of generalization more challenging.

\subsection{Random Initialization}
In addition to our reported results, we additionally attempted to evaluate OpenVLA (Bridge, Rand Init, FT), a model with the same architecture as OpenVLA, but trained from randomly initialized weights, rather than pre-trained VLM weights. Although training metrics on Bridge V2 and our in-domain data did converge, we found that the model failed completely to succeed at our in-distribution base tasks. Therefore, we did not consider this model in our evaluations.

\subsection{Evaluation Conditions}
\label[appendix]{sec:conditions_details}

We detail every generalization condition we consider in our evaluation, including scene images, names for each evaluation condition, axes, language instructions, and additional notes. We provide this information for the main evaluations based on the "put carrot" and "put knife" tasks in \cref{tab:carrot_knife_conditions_1}, for the "flip pot" and "put plate" tasks in \cref{tab:pot_plate_conditions_1}, and for the compositional evaluations in \cref{tab:carrot_knife_compositional}. %

\smallskip \noindent \textbf{Implementation Details}. For the generalization conditions that involve recoloring the sink (Carrot Red Sink, Knife Red Sink, Pot Green Sink, Plate Green Sink), we do not physically change the sink color. Instead, we preprocess each image frame given to the policy using SAM 2.1 Large~\cite{ravi2024sam2} to segment the sink, and then recoloring these pixels to the desired color.

For the generalization conditions that make the sink table shorter relative to the robot (Carrot Shorter Table, Knife Shorter Table, Pot Shorter Table, Plate Shorter Table), rather than lowering the sink table, we raise the table that the robot is mounted on, which achieves the same effect.

{%
\renewcommand{\arraystretch}{2.75}
\begin{table*}[h]
    \fontsize{8}{8}\selectfont
    \centering
    \begin{tabular}{ccccc}
        \toprule
        \textbf{Scene Image} & \textbf{Condition Name} & \textbf{Axis} & \textbf{Language Instruction} & \textbf{Notes} \\
        \midrule
        \multirow{10}{*}{{\includegraphics[width=0.2\textwidth]{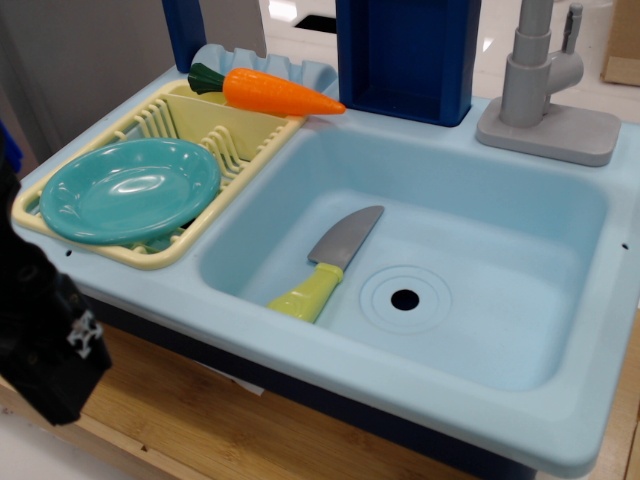}}} & Carrot Base & In-distribution & put carrot on plate & N/A \\ 
         & Knife Base & In-distribution & put knife on plate & N/A  \\
         & Carrot Color & S-PROP & \makecell{put the orange object\\on the plate} & refers to carrot by color \\
         & Knife Color & S-PROP & \makecell{put the gray and green\\object on the plate} & refers to knife by color \\
         & Carrot Lift/Place & S-LANG & lift carrot and place on plate & \makecell{replaced verb ``put"\\using ``lift" and ``place"} \\
         & Knife Lift/Place & S-LANG & lift knife and place on plate & \makecell{replaced verb ``put"\\using ``lift" and ``place"} \\
         & Carrot Counter & S-MO & \makecell{put the object that is\\on the counter on the plate} & \makecell{refers to carrot by\\location (on counter)} \\
         & Knife Sink & S-MO & \makecell{put the object that is\\in the sink on the plate} & \makecell{refers to knife\\by location (in sink)} \\
         & Carrot Basketball & S-INT & \makecell{put the object that is the same\\color as a basketball on the plate} & \makecell{refers to carrot by color\\of a basketball (orange)} \\
         & Knife Typo & S-INT & put knif on plate & ``knife" misspelled as ``knif"  \\
         & Carrot in Sink & SB-SMO & put carrot in sink & \makecell{goal for carrot is\\sink instead of plate} \\
         & Rotate Knife & SB-VRB & rotate knife clockwise & \makecell{rotate knife instead\\of put on plate} \\
        \midrule
        \multirow{2}{*}{{\includegraphics[width=0.2\textwidth]{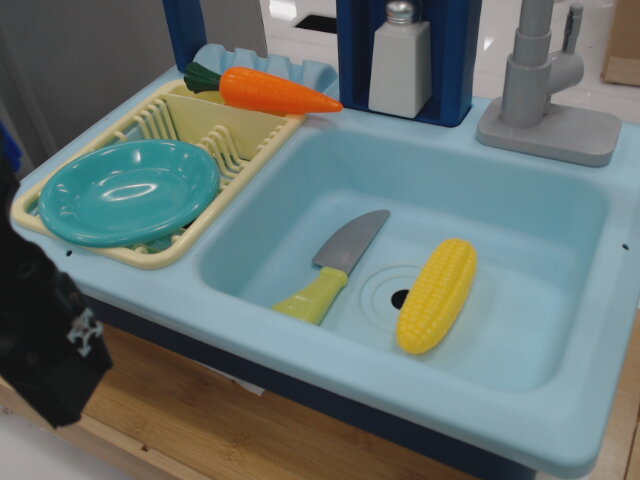}}} & Carrot Distractors & V-SC & put carrot on plate & \makecell{distractor objects \\(corn, salt shaker)} \\
         & Knife Distractors & V-SC & put knife on plate & \makecell{distractor objects\\(corn, salt shaker)}  \\
        \addlinespace[34pt]
        \midrule
        \multirow{2}{*}{{\includegraphics[width=0.2\textwidth]{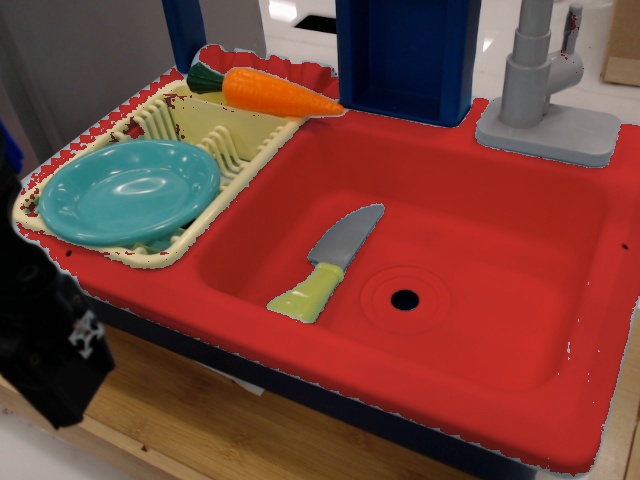}}} 
         & Carrot Red Sink & V-SC & put carrot on plate & red sink  \\
         & Knife Red Sink & V-SC & put knife on plate & red sink  \\
        \addlinespace[34pt]
        \midrule
        \multirow{2}{*}{{\includegraphics[width=0.2\textwidth]{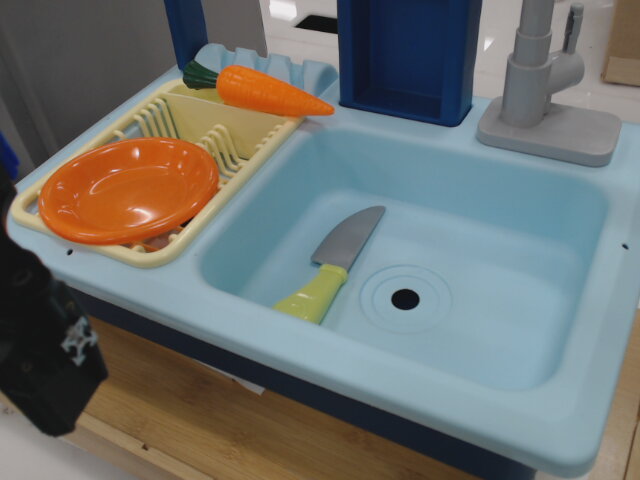}}} 
         & Carrot Orange Plate & V-OBJ & put carrot on plate & orange plate  \\
         & Knife Orange Plate & V-OBJ & put knife on plate & orange plate  \\
        \addlinespace[34pt]
        \midrule
        \multirow{2}{*}{{\includegraphics[width=0.2\textwidth]{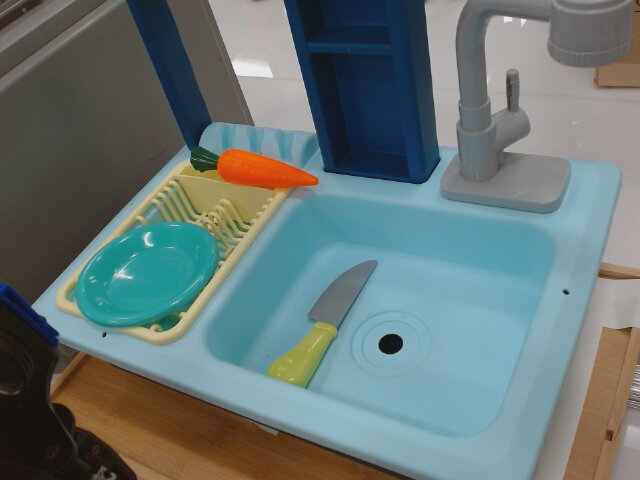}}} 
         & Carrot Camera & V-VIEW & put carrot on plate & new camera pose  \\
         & Knife Camera & V-VIEW & put knife on plate & new camera pose  \\
        \addlinespace[34pt]
        \bottomrule
    \end{tabular}
    \caption{Generalization conditions for ``put carrot" and ``put knife".}
    \label{tab:carrot_knife_conditions_1}
\end{table*}

\begin{table*}[h]
    \ContinuedFloat
    \fontsize{8}{8}\selectfont
    \centering
    \begin{tabular}{ccccc}
        \toprule
        \textbf{Scene Image} & \textbf{Condition Name} & \textbf{Axis} & \textbf{Language Instruction} & \textbf{Notes} \\
        \midrule
        \multirow{1}{*}{\includegraphics[width=0.2\textwidth]{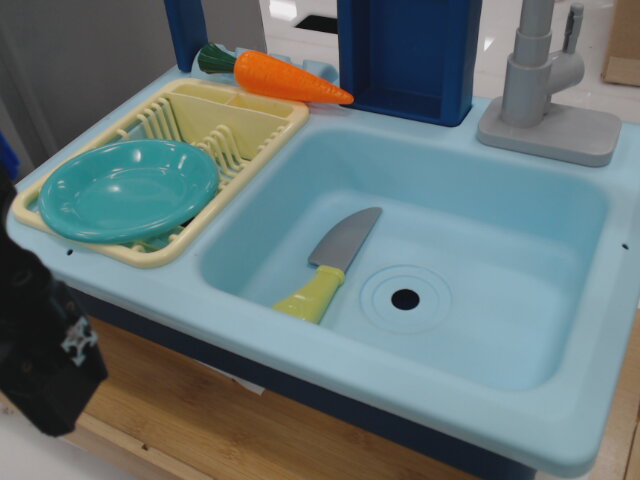}}
         & Carrot Farther & VB-POSE & put carrot on plate & \makecell{carrot slightly farther\\from robot}  \\
        \addlinespace[57pt]
        \midrule
        \multirow{1}{*}{\includegraphics[width=0.2\textwidth]{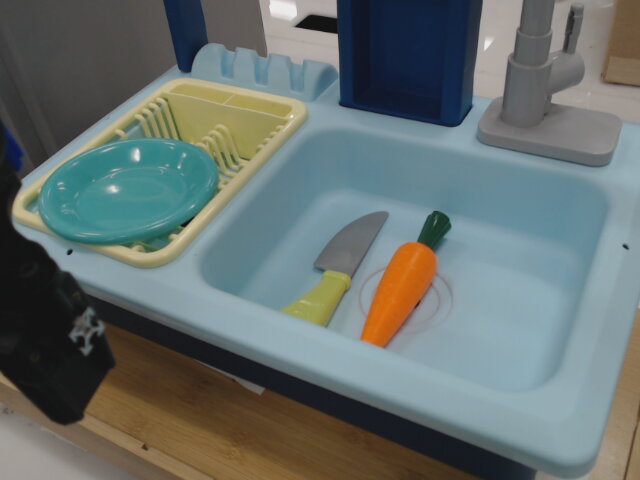}} & Carrot Start Sink & VB-POSE & put carrot on plate & \makecell{carrot start in sink,\\oriented vertically} \\
        \addlinespace[57pt]
        \midrule
        \multirow{1}{*}{\includegraphics[width=0.2\textwidth]{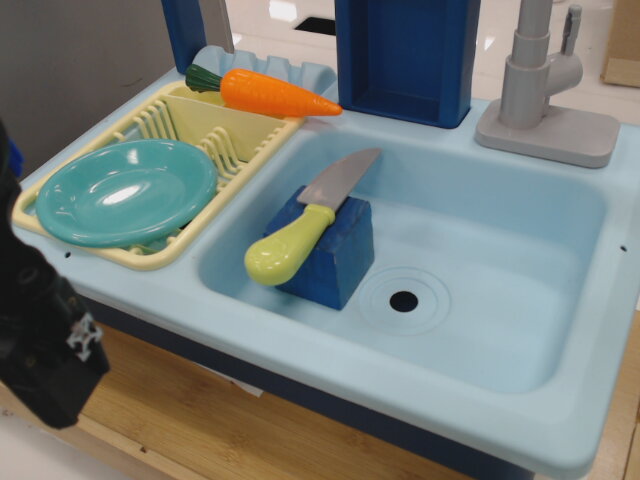}} & Knife Raised & VB-POSE & put knife on plate & \makecell{knife raised by\\placing it on a block} \\
        \addlinespace[57pt]
        \midrule
        \multirow{1}{*}{\includegraphics[width=0.2\textwidth]{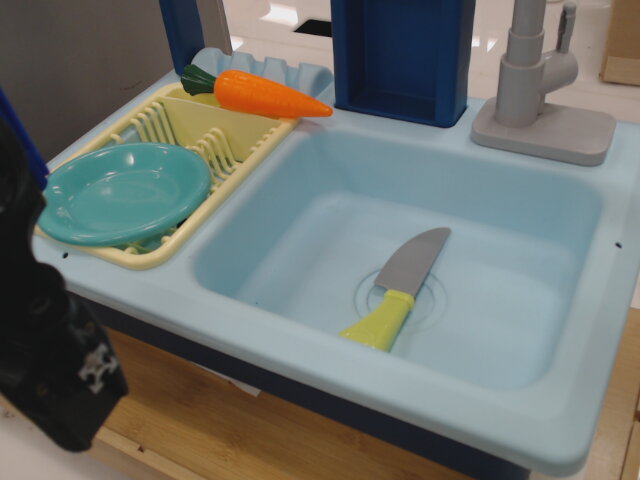}} & Knife Right & VB-POSE & put knife on plate & knife moved to the right \\
        \addlinespace[57pt]
        \midrule
        \multirow{2}{*}{\includegraphics[width=0.2\textwidth]{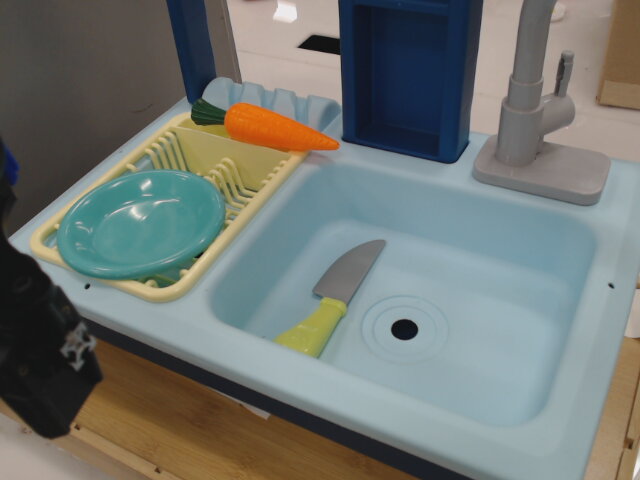}} & Carrot Shorter Table & VB-ISC & put carrot on plate & shorter sink table \\
        & Knife Shorter Table & VB-ISC & put knife on plate & shorter sink table \\
        \addlinespace[34pt]
        \midrule
        \multirow{1}{*}{\includegraphics[width=0.2\textwidth]{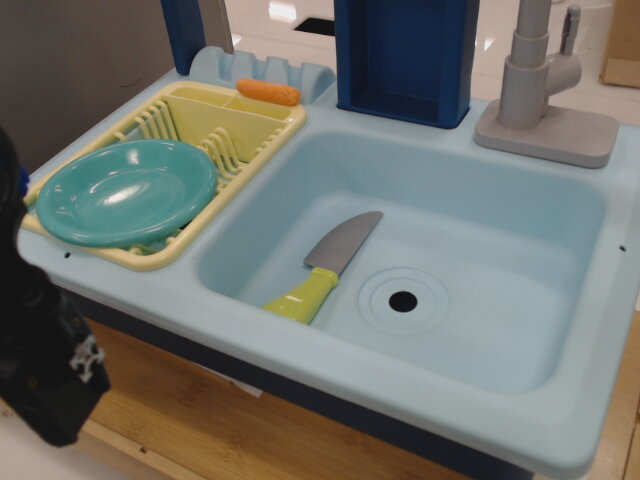}} & Baby Carrot & VB-MOBJ & put carrot on plate & real baby carrot \\
        \addlinespace[57pt]
        \midrule
        \multirow{1}{*}{\includegraphics[width=0.2\textwidth]{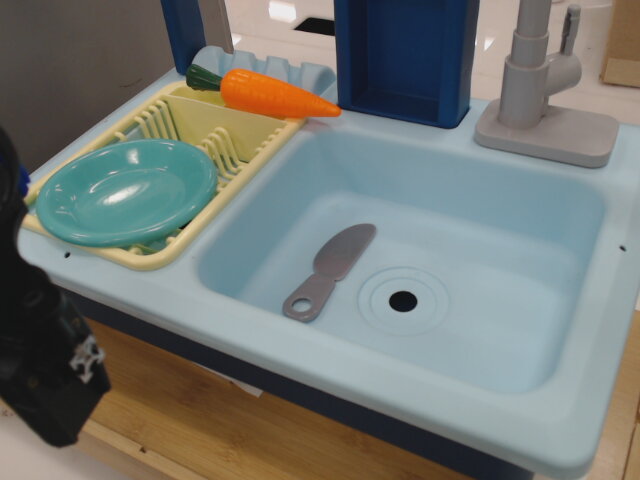}} & Small Knife & VB-MOBJ & put knife on plate & smaller knife \\
        \addlinespace[57pt]
        \bottomrule
    \end{tabular}
    \caption{Generalization conditions for ``put carrot" and ``put knife".}
    \label{tab:carrot_knife_condtiions_2}
\end{table*}

\begin{table*}[h]
    \ContinuedFloat
    \fontsize{8}{8}\selectfont
    \centering
    \begin{tabular}{ccccc}
        \toprule
        \textbf{Scene Image} & \textbf{Condition Name} & \textbf{Axis} & \textbf{Language Instruction} & \textbf{Notes} \\
        \midrule
        \multirow{1}{*}{\includegraphics[width=0.2\textwidth]{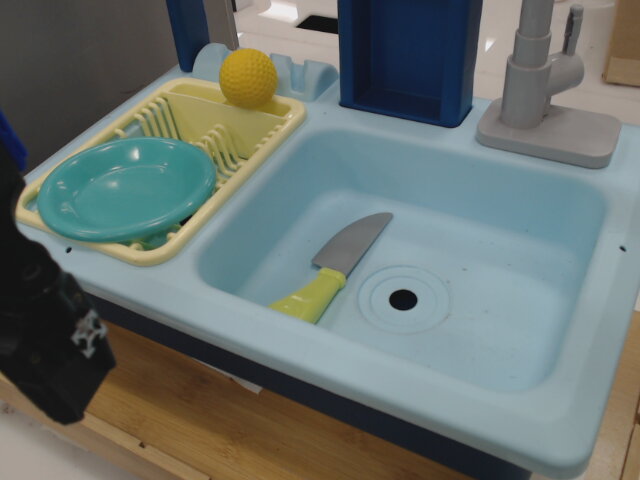}} & Ball & VSB-NOBJ & put ball on plate & carrot replaced with ball \\
        \addlinespace[57pt]
        \midrule
        \multirow{1}{*}{\includegraphics[width=0.2\textwidth]{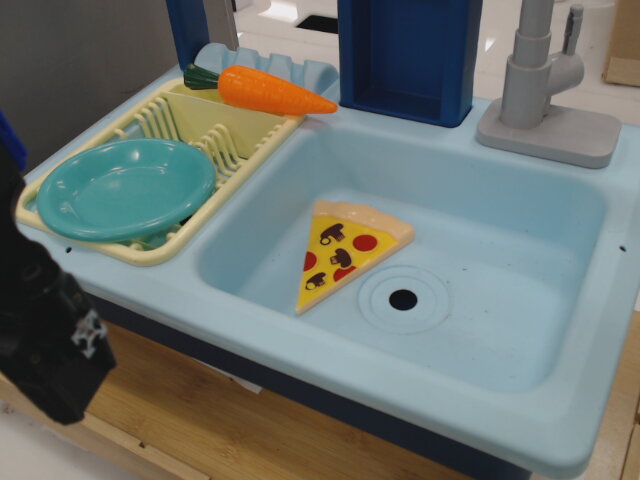}} & Pizza & VSB-NOBJ & put pizza on plate & knife replaced with pizza \\
        \addlinespace[57pt]
        \bottomrule
    \end{tabular}
    \caption{Generalization conditions for ``put carrot" and ``put knife".}
    \label{tab:carrot_knife_condtiions_3}
\end{table*}

\begin{table*}[h]
    \fontsize{8}{8}\selectfont
    \centering
    \begin{tabular}{ccccc}
        \toprule
        \textbf{Scene Image} & \textbf{Condition Name} & \textbf{Axis} & \textbf{Language Instruction} & \textbf{Notes} \\
        \midrule
        \multirow{10}{*}{{\includegraphics[width=0.2\textwidth]{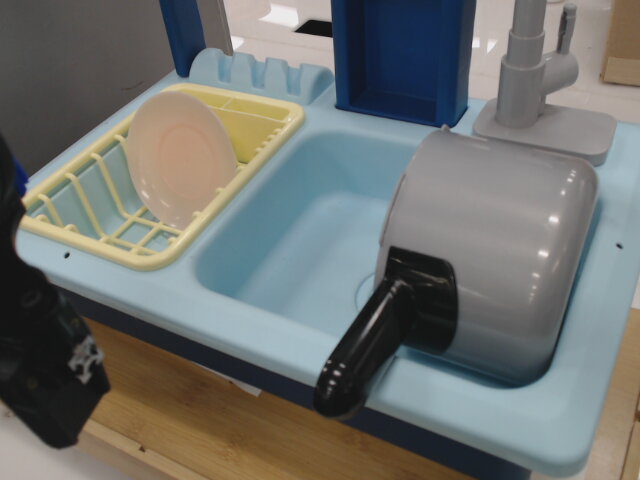}}} & Pot Base & In-distribution & \makecell{flip pot upright\\which is in sink} & N/A \\ 
         & Plate Base & In-distribution & put plate in sink & N/A  \\
         & Pot Color & S-PROP & \makecell{flip the gray object upright\\which is in sink} & refers to pot by color \\
         & Plate Color & S-PROP & \makecell{put the pink object\\in the sink} & refers to plate by color \\
         & Pot Lift/Place & S-LANG & \makecell{lift pot upright\\and place in sink} & \makecell{replaced verb ``flip"\\using ``lift" and ``place"} \\
         & Plate Lift/Place & S-LANG & lift plate and place in sink & \makecell{replaced verb ``put"\\using ``lift" and ``place"} \\
         & Pot Sink & S-MO & \makecell{flip the object that  is\\in the sink upright} & \makecell{refers to pot by\\location (in sink)} \\
         & Plate Drying Rack & S-MO & \makecell{put the object that is in\\the drying rack in the sink} & \makecell{refers to plate by\\location (in drying rack)} \\
         & Pot Boiling & S-INT & \makecell{flip the object that can be\\used for boiling water upright} & \makecell{refers to pot by ability\\to boil water} \\
         & Plate Typo & S-INT & put plait on plate & ``plate" misspelled as ``plait"  \\
         & Plate to Counter & SB-SMO & put plate on counter & \makecell{goal for plate is\\counter instead of sink}\\
         & Pot to Left & SB-VRB & \makecell{move pot to the\\left side of the sink} & \makecell{move pot left instead\\of flip upright} \\
        \midrule
        \multirow{2}{*}{{\includegraphics[width=0.2\textwidth]{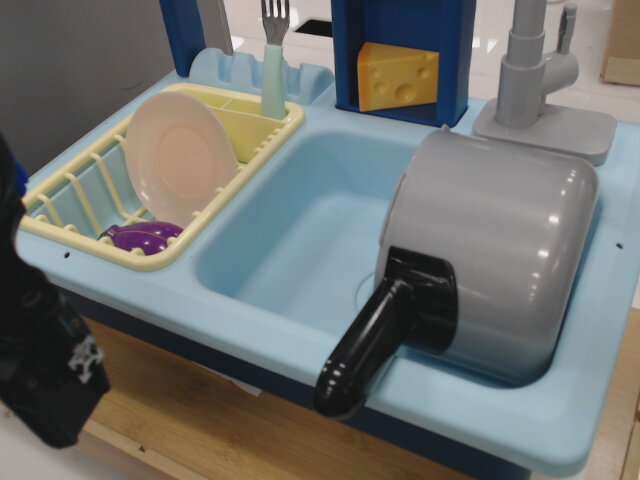}}} & Pot Distractors & V-SC & \makecell{flip pot upright\\which is in sink} & \makecell{distractor objects \\(eggplant, fork, cheese)} \\
         & Plate Distractors & V-SC & put plate in sink & \makecell{distractor objects\\(eggplant, fork, cheese)}  \\
        \addlinespace[34pt]
        \midrule
        \multirow{2}{*}{{\includegraphics[width=0.2\textwidth]{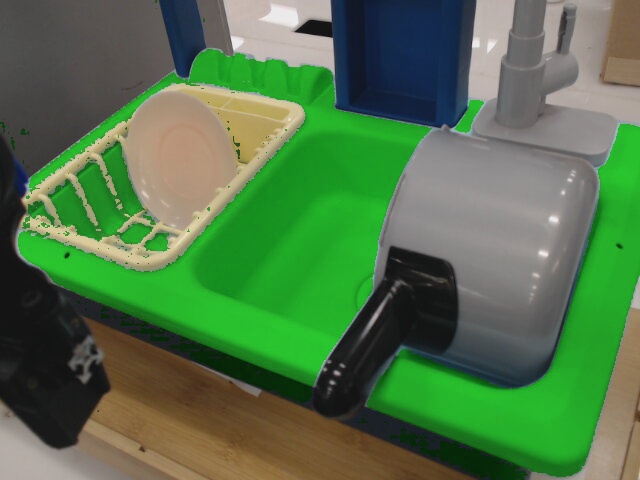}}} 
         & Pot Green Sink & V-SC & \makecell{flip pot upright\\which is in sink} & green sink  \\
         & Plate Green Sink & V-SC & put plate in sink & green sink  \\
        \addlinespace[34pt]
        \midrule
        \multirow{1}{*}{{\includegraphics[width=0.2\textwidth]{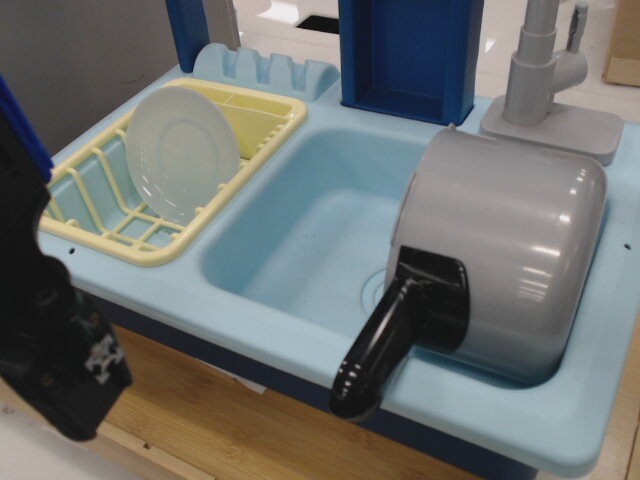}}} 
         & Gray Plate & V-OBJ & put plate in sink & gray plate  \\
        \addlinespace[57pt]
        \midrule
        \multirow{2}{*}{{\includegraphics[width=0.2\textwidth]{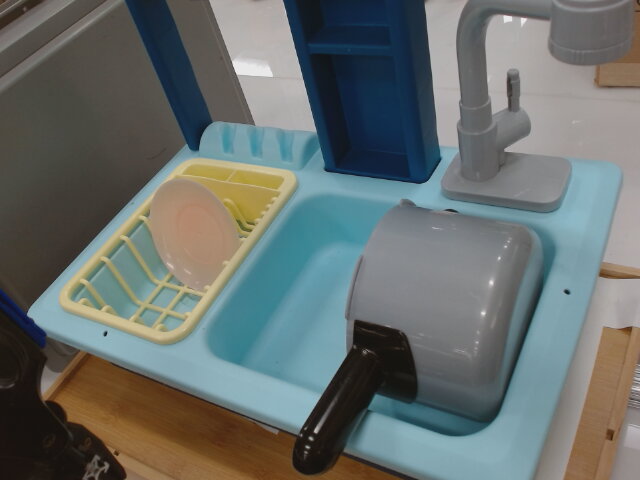}}} 
         & Pot Camera & V-VIEW & \makecell{flip pot upright\\which is in sink} & new camera pose  \\
         & Plate Camera & V-VIEW & put plate in sink & new camera pose  \\
        \addlinespace[34pt]
        \bottomrule
    \end{tabular}
    \caption{Generalization conditions for ``flip pot" and ``put plate".}
    \label{tab:pot_plate_conditions_1}
\end{table*}

\begin{table*}[h]
    \ContinuedFloat
    \fontsize{8}{8}\selectfont
    \centering
    \begin{tabular}{ccccc}
        \toprule
        \textbf{Scene Image} & \textbf{Condition Name} & \textbf{Axis} & \textbf{Language Instruction} & \textbf{Notes} \\
        \midrule
        \multirow{1}{*}{\includegraphics[width=0.2\textwidth]{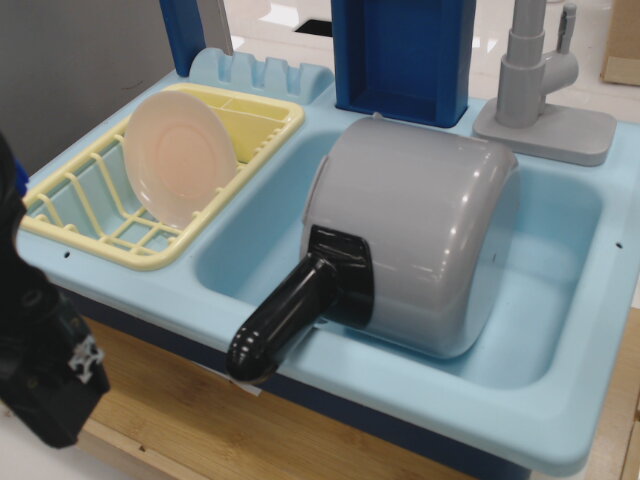}}
         & Pot Left & VB-POSE & \makecell{flip pot upright\\which is in sink} & pot moved to left  \\
        \addlinespace[57pt]
        \midrule
        \multirow{1}{*}{\includegraphics[width=0.2\textwidth]{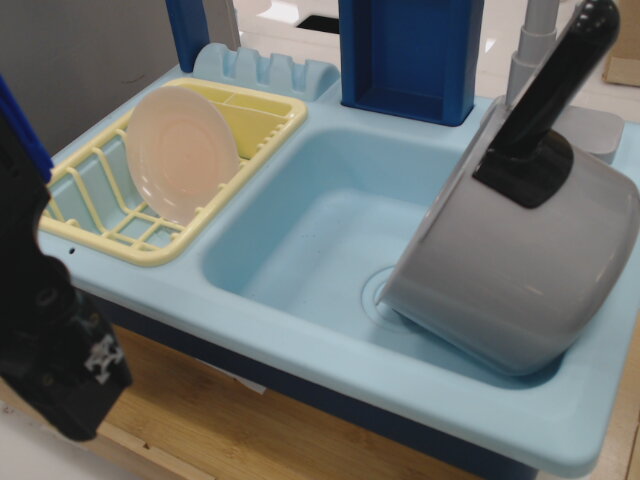}} & Pot Angled & VB-POSE & \makecell{flip pot upright\\which is in sink} & \makecell{pot rotated and\\angled to the right} \\
        \addlinespace[57pt]
        \midrule
        \multirow{1}{*}{\includegraphics[width=0.2\textwidth]{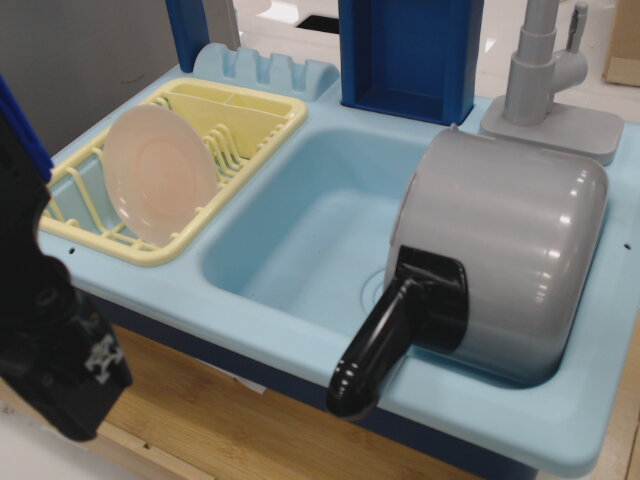}} & Plate Closer & VB-POSE & put plate in sink & \makecell{plate slightly\\closer to robot} \\
        \addlinespace[57pt]
        \midrule
        \multirow{1}{*}{\includegraphics[width=0.2\textwidth]{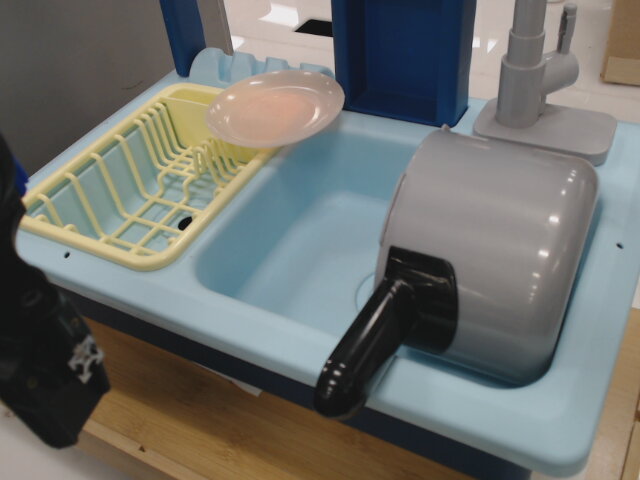}} & Plate Counter & VB-POSE & put plate in sink & plate flat on counter \\
        \addlinespace[57pt]
        \midrule
        \multirow{2}{*}{\includegraphics[width=0.2\textwidth]{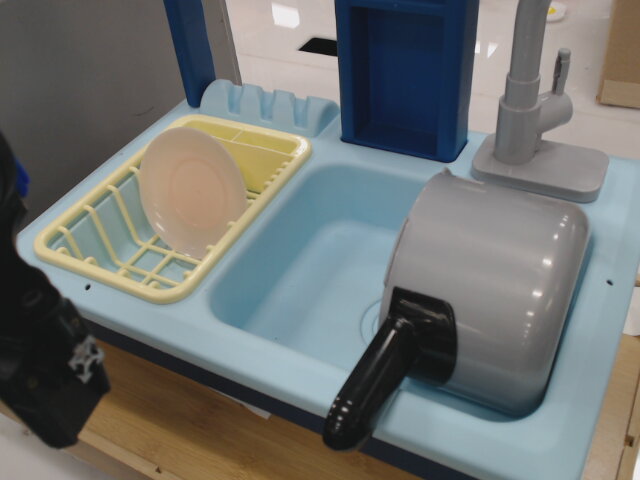}} & Pot Shorter Table & VB-ISC & \makecell{flip pot upright\\which is in sink} & shorter sink table \\
        & Plate Shorter Table & VB-ISC & put plate in sink & shorter sink table \\
        \addlinespace[34pt]
        \midrule
        \multirow{1}{*}{\includegraphics[width=0.2\textwidth]{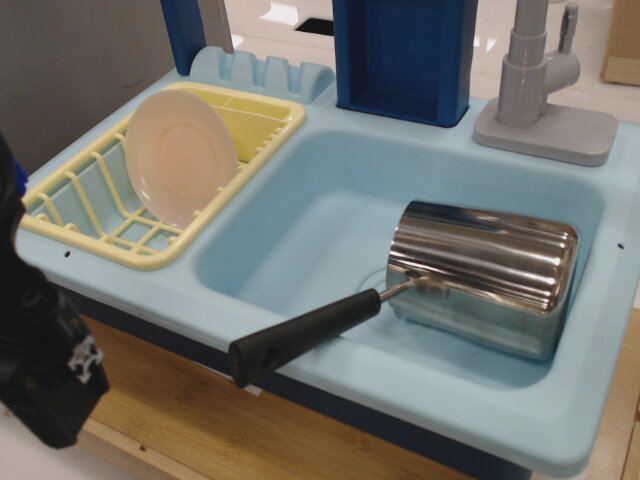}} & Thin Pot  & VB-MOBJ & \makecell{flip pot upright\\which is in sink} & \makecell{thinner and taller\\metal pot} \\
        \addlinespace[57pt]
        \midrule
        \multirow{1}{*}{\includegraphics[width=0.2\textwidth]{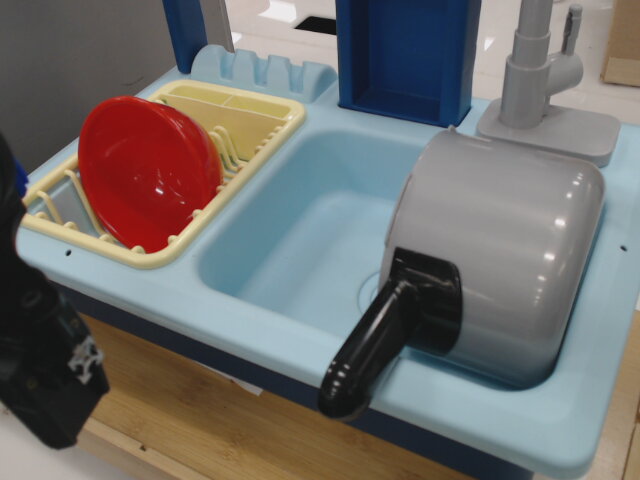}} & Red Bowl & VB-MOBJ & put plate in sink & \makecell{plate replaced\\with red bowl} \\
        \addlinespace[57pt]
        \bottomrule
    \end{tabular}
    \caption{Generalization conditions for ``flip pot" and ``put plate".}
    \label{tab:pot_plate_condtiions_2}
\end{table*}

\begin{table*}[h]
    \ContinuedFloat
    \fontsize{8}{8}\selectfont
    \centering
    \begin{tabular}{ccccc}
        \toprule
        \textbf{Scene Image} & \textbf{Condition Name} & \textbf{Axis} & \textbf{Language Instruction} & \textbf{Notes} \\
        \midrule
        \multirow{1}{*}{\includegraphics[width=0.2\textwidth]{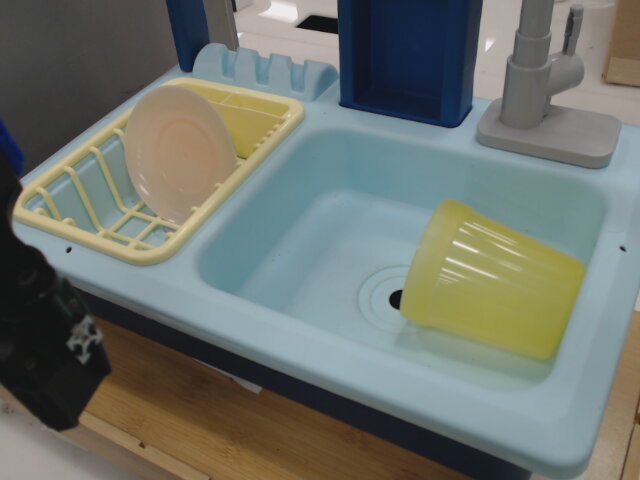}} & Cup & VSB-NOBJ & \makecell{flip cup upright\\which is in sink} & pot replaced with cup \\
        \addlinespace[57pt]
        \midrule
        \multirow{1}{*}{\includegraphics[width=0.2\textwidth]{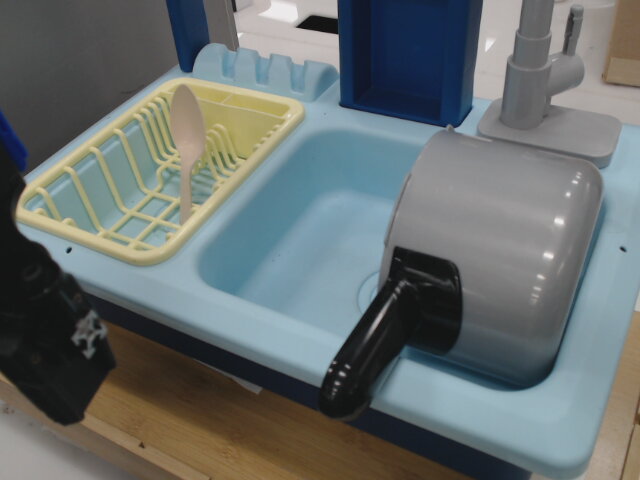}}
         & Spoon & VSB-NOBJ  & \makecell{flip pot upright\\which is in sink} & plate replaced with spoon  \\
        \addlinespace[57pt]
        \bottomrule
    \end{tabular}
    \caption{Generalization conditions for ``flip pot" and ``put plate".}
    \label{tab:pot_plate_condtiions_3}
\end{table*}

\begin{table*}[h]
    \fontsize{8}{8}\selectfont
    \centering
    \begin{tabular}{ccccc}
        \toprule
        \textbf{Scene Image} & \textbf{Condition Name} & \textbf{Axis} & \textbf{Language Instruction} & \textbf{Notes} \\
        \midrule
        \multirow{2}{*}{\includegraphics[width=0.2\textwidth]{figures/scenes/carrot-knife/base.jpg}}
         & \makecell{Carrot Color +\\Lift/Place} & S-PROP + S-LANG & \makecell{lift the orange object\\and place on plate} & \makecell{refers to carrot by color\\and replaced verb ``put"\\using ``lift" and ``place"}  \\
         & \makecell{Knife Color +\\Lift/Place} & S-PROP + S-LANG & \makecell{lift the gray and green\\object and place on plate} & \makecell{refers to knife by color\\and replaced verb ``put"\\using ``lift" and ``place"}  \\
        \addlinespace[28pt]
        \midrule
        \multirow{2}{*}{\includegraphics[width=0.2\textwidth]{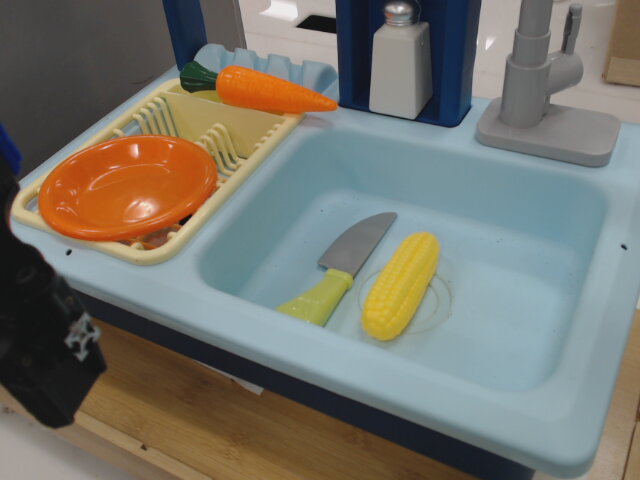}} & \makecell{Carrot Distractors +\\Orange Plate} & V-SC + V-OBJ & put carrot on plate & \makecell{distractor objects\\(corn, salt shaker)\\and orange plate} \\
         & \makecell{Knife Distractors +\\Orange Plate} & V-SC + V-OBJ & put knife on plate & \makecell{distractor objects\\(corn, salt shaker)\\and orange plate} \\
        \addlinespace[28pt]
        \midrule
        \multirow{1}{*}{\includegraphics[width=0.2\textwidth]{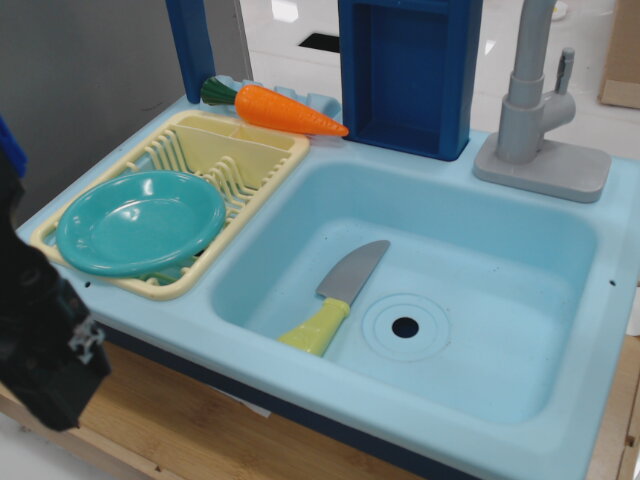}} & \makecell{Carrot Farther +\\Shorter Table} & VB-POSE + VB-ISC & put carrot on plate & \makecell{carrot slightly farther\\from robot and\\shorter sink table} \\
        \addlinespace[53pt]
        \midrule
        \multirow{1}{*}{\includegraphics[width=0.2\textwidth]{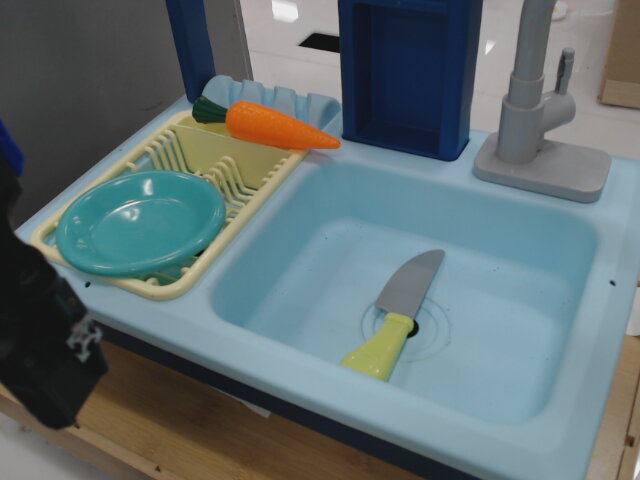}} & \makecell{Knife Right +\\Shorter Table} & VB-POSE + VB-ISC & put knife on plate & \makecell{knife moved to right and\\shorter sink table} \\
        \addlinespace[57pt]
        \bottomrule
    \end{tabular}
    \caption{Generalization conditions for compositional experiments.}
    \label{tab:carrot_knife_compositional}
\end{table*}
}

\clearpage

\subsection{More Evaluation Results}
\subsubsection{Composition}
\label[appendix]{sec:compositional_results}

\begin{table*}[h]
    \fontsize{7}{8}\selectfont
    \centering
    \begin{tabular}{p{2cm}*7{>{\centering\arraybackslash}p{2cm}}}
    \toprule
                                                        & \makecell{OpenVLA\\(OXE)} & \makecell{OpenVLA\\(OXE, FT)} & \makecell{OpenVLA\\(Bridge, FT)} & \makecell{OpenVLA\\(Bridge, VQA, FT)} & \makecell{MiniVLA\\(Bridge, FT)} & \makecell{MiniVLA\\(Bridge, --VQ, FT)} & \makecell{$\pi_0$ Reimplement \\(Bridge, FT)} \\
    \midrule
    \makecell[l]{\textgbf{Semantic}\\(\SPropShort + \SRephraseShort)}  &  6/10               &  8/10         &  4/10     & 0/10        &  4/10            &  2/10 & 1/10                  \\
    \makecell[l]{\textgbf{Visual}\\(\VSceneShort + \VObjectShort)}      &  3/10               &  6/10         &  5/10    & 6/10         &  8/10            &  5/10 & 7/10                 \\
    \addlinespace[5pt]
    \makecell[l]{\textgbf{Visual + Behavioral}\\(\VBPoseShort + \VBSceneShort)}      &  5/10               &  6/10         &  3/10      & 7/10       &  5/10            &  6/10 & 8/10                  \\
    \midrule
    Overall                                         & 14/30               & 20/30         & 12/30   & 13/30          &  17/30           & 13/30    & 16/30               \\
    \bottomrule
    \end{tabular}
    \caption{Compositional results for two axes from each of \textgbf{semantic}, \textgbf{visual}, and \textgbf{visual + behavioral}.}
    \label{tab:compositional_results}
    \vspace{-5pt}
\end{table*}

We conduct additional evaluations on compositions of our \cref{tab:compositional_results}. We once again see that training on the larger OXE mixture seems to improve many \textgbf{semantic} axes, specifically the composition of referring to object properties in language and rephrasing language instructions (\SPropShort + \SRephraseShort), possibly due to a larger variety of language instructions in the training data. Furthermore, some models are fairly robust to combinations of distractors and object colors (\VSceneShort + \VObjectShort), with MiniVLA and $\pi_0$ being the most performant, similarly as in the main results. Some models are also surprisingly robust to the composition of different object poses and scene factors (\VBPoseShort + \VBSceneShort), with $\pi_0$ performing the best in this setting.

\subsubsection{Detailed Results}
\label[appendix]{sec:results:details}

We provide individual success rates for all generalization conditions listed in \cref{sec:conditions_details}. We provide this for the main evaluations based on the ``put carrot" and ``put knife" tasks in \cref{tab:carrot_knife_results_details}, for the ``flip pot" and ``put plate" tasks in \cref{tab:pot_plate_results_details}, and for the compositional evaluations in \cref{tab:compositional_results_details}.

{%
\renewcommand{\arraystretch}{1.1}
\begin{table*}[hbt!]
    \fontsize{7}{8}\selectfont
    \centering
    \begin{tabular}{cccccccccc}
         \toprule
         \makecell{Condition\\Name} & \makecell{OpenVLA\\(OXE)} & \makecell{OpenVLA\\(OXE, FT)} & \makecell{OpenVLA\\(Bridge, FT)} & \makecell{OpenVLA\\(Bridge, VQA, FT)} & \makecell{MiniVLA\\(Bridge, FT)} & \makecell{MiniVLA\\(Bridge, FT, -VQ)} & \makecell{$\pi_0$ Reimplment\\(Bridge, FT)} \\
         \midrule
         Carrot Base & 3/5 & 5/5 & 3/5 & 4/5 & 5/5 & 4/5 & 4/5 \\
         Knife Base & 2/5 & 3/5 & 5/5 & 5/5 & 5/5 & 4/5 & 5/5 \\
         Carrot Color & 2/5 & 4/5 & 4/5 & 2/5 & 2/5 & 2/5 & 2/5 \\
         Knife Color & 2/5 & 3/5 & 0/5 & 0/5 & 0/5 & 0/5 & 0/5 \\
         Carrot Lift/Place & 3/5 & 5/5 & 2/5 & 4/5 & 5/5 & 3/5 & 5/5 \\
         Knife Lift/Place & 2/5 & 4/5 & 4/5 & 3/5 & 4/5 & 1/5 & 5/5 \\
         Carrot Counter & 0/5 & 0/5 & 0/5 & 0/5 & 0/5 & 0/5 & 0/5 \\
         Knife Sink & 1/5 & 2/5 & 1/5 & 3/5 & 0/5 & 0/5 & 3/5 \\
         Carrot Basketball & 0/5 & 0/5 & 0/5 & 1/5 & 0/5 & 0/5 & 0/5 \\
         Knife Typo & 1/5 & 4/5 & 4/5 & 4/5 & 1/5 & 1/5 & 4/5 \\
         Carrot in Sink & -- & -- & 5/5 & -- & 5/5 & -- & 3/5 \\
         Rotate Knife & -- & -- & 1/5 & -- & 0/5 & -- & 0/5 \\
         Carrot Distractors & 3/5 & 5/5 & 3/5 & 3/5 & 3/5 & 2/5 & 4/5 \\
         Knife Distractors & 1/5 & 3/5 & 2/5 & 3/5 & 5/5 & 3/5 & 4/5 \\
         Carrot Red Sink & 3/5 & 3/5 & 1/5 & 3/5 & 3/5 & 1/5 & 5/5 \\
         Knife Red Sink & 1/5 & 3/5 & 2/5 & 2/5 & 3/5 & 4/5 & 2/5 \\
         Carrot Orange Plate & 2/5 & 3/5 & 2/5 & 3/5 & 4/5 & 0/5 & 3/5 \\
         Knife Orange Plate & 4/5 & 4/5 & 2/5 & 4/5 & 5/5 & 5/5 & 5/5 \\
         Carrot Camera & 0/5 & 0/5 & 0/5 & 0/5 & 1/5 & 0/5 & 0/5 \\
         Knife Camera & 0/5 & 0/5 & 0/5 & 0/5 & 1/5 & 0/5 & 5/5 \\
         Carrot Farther & 2/5 & 3/5 & 2/5 & 4/5 & 3/5 & 3/5 & 2/5 \\
         Carrot Start Sink & 3/5 & 3/5 & 3/5 & 2/5 & 5/5 & 3/5 & 5/5 \\
         Knife Raised & 0/5 & 2/5 & 3/5 & 4/5 & 4/5 & 3/5 & 1/5 \\
         Knife Right & 2/5 & 3/5 & 3/5 & 5/5 & 4/5 & 2/5 & 5/5 \\
         Carrot Shorter Table & 2/5 & 5/5 & 2/5 & 4/5 & 3/5 & 4/5 & 5/5 \\
         Knife Shorter Table & 0/5 & 3/5 & 3/5 & 3/5 & 2/5 & 2/5 & 5/5 \\
         Baby Carrot & 0/5 & 0/5 & 0/5 & 0/5 & 1/5 & 0/5 & 1/5 \\
         Small Knife & 0/5 & 0/5 & 1/5 & 0/5 & 0/5 & 0/5 & 0/5 \\
         Ball & -- & -- & 0/5 & -- & 3/5 & -- & 1/5 \\
         Pizza & -- & -- & 1/5 & -- & 0/5 & -- & 0/5 \\
         \bottomrule
    \end{tabular}
    \caption{Detailed generalization results for ``put carrot" and ``put knife".}
    \label{tab:carrot_knife_results_details}
\end{table*}

\begin{table*}[]
    \centering
    \fontsize{7}{8}\selectfont
    \begin{tabular}{cccccccccc}
         \toprule
         \makecell{Condition\\Name} & \makecell{OpenVLA\\(Bridge, FT)} & \makecell{MiniVLA\\(Bridge, FT)} & \makecell{$\pi_0$ Reimplment\\(Bridge, FT)} \\
         \midrule
         Pot Base & 3/5 & 5/5 & 5/5 \\
         Plate Base & 3/5 & 4/5 & 4/5 \\
         Pot Color & 4/5 & 1/5 & 1/5 \\
         Plate Color & 0/5 & 0/5 & 0/5 \\
         Pot Lift/Place & 0/5 & 2/5 & 1/5 \\
         Plate Lift/Place & 0/5 & 0/5 & 0/5 \\
         Pot Sink & 1/5 & 1/5 & 5/5 \\
         Plate Drying Rack & 0/5 & 0/5 & 0/5 \\
         Pot Boiling & 0/5 & 0/5 & 0/5 \\
         Plate Typo & 0/5 & 0/5 & 0/5 \\
         Plate to Counter & 0/5 & 0/5 & 0/5 \\
         Pot to Left & 0/5 & 2/5 & 0/5  \\
         Pot Distractors & 3/5 & 4/5 & 5/5 \\
         Plate Distractors & 3/5 & 1/5 & 0/5 \\
         Pot Green Sink & 3/5 & 3/5 & 4/5 \\
         Plate Green Sink & 2/5 & 1/5 & 3/5 \\
         Gray Plate & 4/5 & 3/5 & 3/5 \\
         Pot Camera & 0/5 & 1/5 & 0/5 \\
         Plate Camera & 0/5 & 0/5 & 0/5 \\
         Pot Left & 3/5 & 0/5 & 0/5 \\
         Pot Angled & 3/5 & 3/5 & 5/5 \\
         Plate Closer & 0/5 & 3/5 & 4/5 \\
         Plate Counter & 2/5 & 0/5 & 0/5 \\
         Pot Shorter Table & 5/5 & 2/5 & 5/5 \\
         Plate Shorter Table & 3/5 & 0/5 & 2/5 \\
         Thin Pot & 3/5 & 0/5 & 1/5 \\
         Red Bowl & 0/5 & 2/5 & 5/5 \\
         Cup & 3/5 & 0/5 & 3/5 \\
         Spoon & 0/5 & 0/5 & 0/5 \\
         \bottomrule
    \end{tabular}
    \caption{Detailed generalization results for ``flip pot" and ``put plate".}
    \label{tab:pot_plate_results_details}
\end{table*}

\begin{table*}[]
    \centering
    \fontsize{7}{8}\selectfont
    \begin{tabular}{cccccccccc}
         \toprule
         \makecell{Condition\\Name} & \makecell{OpenVLA\\(OXE)} & \makecell{OpenVLA\\(OXE, FT)} & \makecell{OpenVLA\\(Bridge, FT)} & \makecell{OpenVLA\\(Bridge, VQA, FT)} & \makecell{MiniVLA\\(Bridge, FT)} & \makecell{MiniVLA\\(Bridge, FT, No VQ)} & \makecell{$\pi_0$ Reimplment\\(Bridge, FT)} \\
         \midrule
         \makecell{Carrot Color +\\Lift/Place} & 3/5 & 4/5 & 4/5 & 0/5 & 4/5 & 2/5 & 1/5 \\
         \makecell{Knife Color +\\Lift/Place} & 3/5 & 4/5 & 0/5 & 0/5 & 0/5 & 0/5 & 0/5 \\
         \makecell{Carrot Distractors +\\Orange Plate} & 1/5 & 4/5 & 2/5 & 3/5 & 3/5 & 0/5 & 3/5 \\
         \makecell{Knife Distractors +\\Orange Plate} & 2/5 & 2/5 & 3/5 & 3/5 & 5/5 & 5/5 & 4/5 \\
         \makecell{Carrot Farther +\\Shorter Table} & 3/5 & 3/5 & 3/5 & 3/5 & 2/5 & 3/5 & 3/5 \\
         \makecell{Knife Right +\\Shorter Table} & 2/5 & 3/5 & 0/5 & 4/5 & 3/5 & 3/5 & 5/5 \\
         \bottomrule
    \end{tabular}
    \caption{Detailed results for our compositional evaluation.}
    \label{tab:compositional_results_details}
\end{table*}
}

\clearpage

\begin{figure*}[t]
    \centering
    \includegraphics[width=0.7\linewidth]{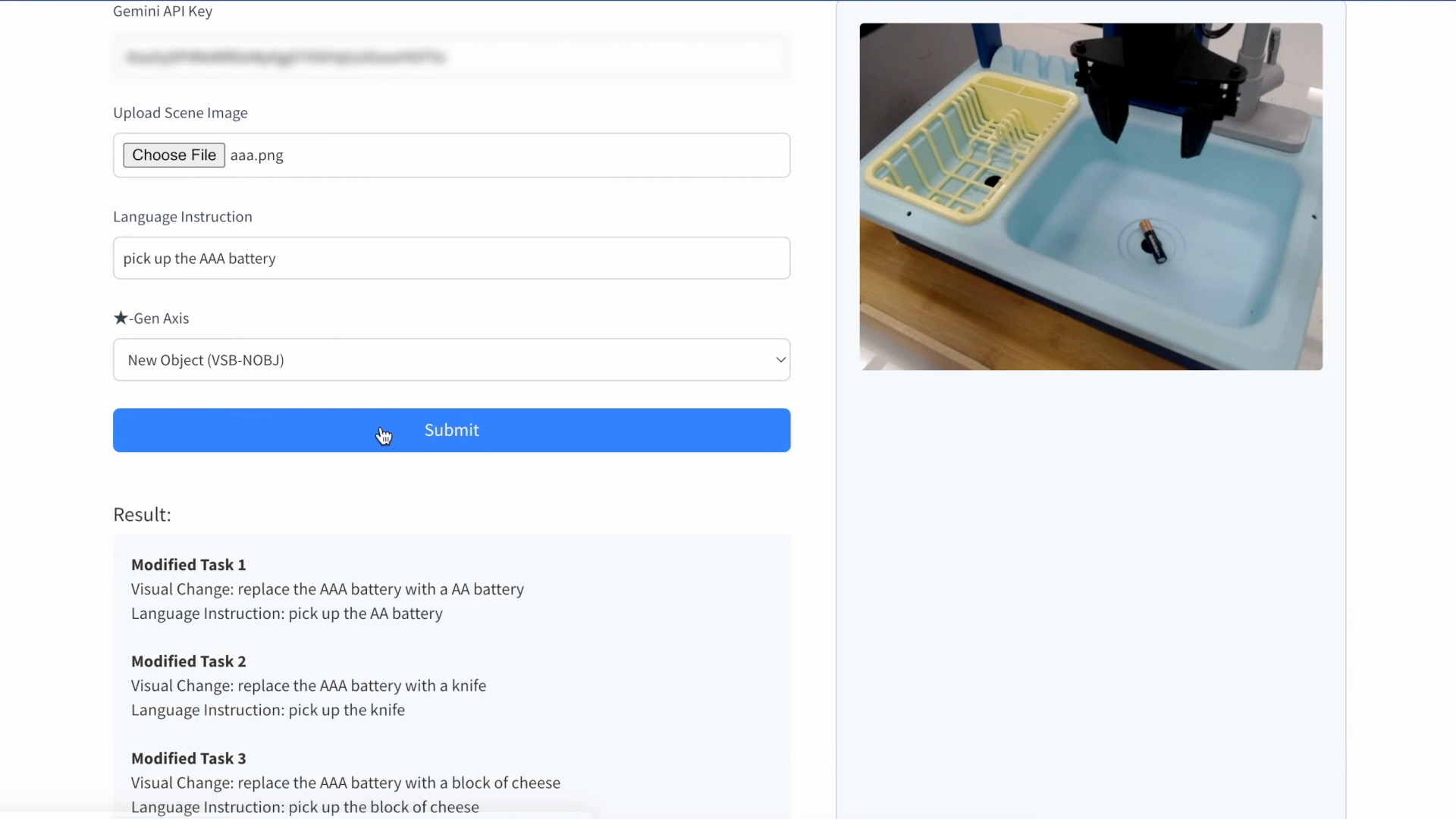}
    \caption{Example of using our demonstration to generate \emph{\VSBObj} perturbations for the base task ``pick up the AAA battery".}
    \vspace{-10pt}
    \label{fig:gen_example}
\end{figure*}

\section{Automatic Benchmark Design}
\label[appendix]{sec:auto_benchmark}

We intend for \TaxonomyName to provide guidance for more extensive and fine-grained evaluations of generalist robot policies. However, we acknowledge that the evaluation conditions considered in \BenchName, while based on \TaxonomyName, were still chosen using human oversight. We believe that it would be beneficial to limit human involvement when constructing future evaluations, because it can require a significant amount of human effort to design a thorough set of evaluation conditions, and human involvement can result in bias, even if unintentional.

Therefore, we make a preliminary step towards reducing human involvement in robot benchmark design by using foundation models to automatically propose evaluation conditions. In particular, we condition a vision-language model (VLM) on a base task, and ask it to propose perturbations of the base task to evaluate a given axis of generalization from \TaxonomyName.

To do this, we prompt the VLM with an initial scene image for the base task, as well as a text prompt that consists of the base task language instruction, and directions for how to modify the task according to a given axis from \TaxonomyName. %
If the axis is \textgbf{visual}, we ask the VLM to specify the perturbation as an instruction for an image-editing model that would perform the perturbation. If the axis is \textgbf{semantic}, we ask for the VLM to generate new language instructions for the perturbed tasks. The VLM can generate both image edits and new language instructions if the given axis is both \textgbf{visual} and \textgbf{semantic}.

\begin{figure}
\begin{tcolorbox}[colback=gray!5, colframe=gray!75]
This is an image of a scene where a robot is to complete the task ``put carrot on plate". Suggest 3 changes to the task that each involve changing a single task-relevant object to a new object with a different visual appearance, semantic description, and physical characteristics. Do this by providing 3 updated language instructions for each of the modified tasks, and corresponding text prompts to an image-editing model that would each perform a single change to the scene to create the modified task. Remember to only change one object, and to only change an object that is involved in the task.
\end{tcolorbox}
\vspace{-5pt}
\caption{Example text prompt for generating perturbations of the base task ``put carrot on plate" for the axis \emph{\VSBObj}.}
\vspace{-7.5pt}
\label{fig:vlm_prompt}
\end{figure}

We developed an interactive demonstration of this system where users can provide their own base tasks (as initial scene images and language instructions) on our \hyperlink{stargen-taxonomy.github.io}{website}. We use Gemini 2.0 Flash as our VLM. For more reliable generations, we configure Gemini to generate perturbed tasks as JSON, using a specified schema. We instantiate the system for the following 5 \TaxonomyName axes:

\begin{enumerate}
    \item \emph{\VObject} (\VObjectShort)
    \item \emph{\SProp} (\SPropShort)
    \item \emph{\VBPose} (\VBPoseShort)
    \item \emph{\SBAction} (\SBActionShort)
    \item \emph{\VSBObj} (\VSBObjShort)
\end{enumerate}

\noindent In \cref{fig:gen_example}, we provide an example screenshot of the demonstration used to generate perturbations for a base task that was not used in \BenchName. In \cref{fig:vlm_prompt}, we provide an example text prompt used for the axis \emph{\VSBObj}, and in \cref{fig:json_schema} we provide the Node.js JSON schema used for this axis.

\begin{figure}
\begin{tiny}
\begin{verbatim}

import { SchemaType } from "@google/generative-ai";

const visualSemanticSchema = {
    description: "Visual and language instruction changes for a task.",
    type: SchemaType.ARRAY,
    items: {
      type: SchemaType.OBJECT,
      properties: {
        visualChange: {
          type: SchemaType.STRING,
          description: "Text prompt for an image-editing model to modify the task",
          nullable: false,
        },
        languageChange: {
          type: SchemaType.STRING,
          description: "Updated language instruction for the modified task",
          nullable: false,
        },
      },
      required: ["visualChange", "languageChange"],
    },
};
\end{verbatim}
\end{tiny}
\caption{Node.js JSON schema used for Gemini API to generate perturbations that are both \textgbf{visual} and \textgbf{semantic}, such as those for the axis \emph{\VSBObj}.}
\vspace{-12.5pt}
\label{fig:json_schema}
\end{figure}

\end{document}